%% file: main.tex
\documentclass[sigconf,nonacm,screen]{acmart}

\setcitestyle{round}
\bibpunct{(}{)}{;}{a}{,}{,}

\usepackage{enumitem}
\setlist{leftmargin=5.5mm}

\usepackage[skip=3pt]{caption}
\usepackage{subcaption}
\usepackage{multirow}
\usepackage{array}
\usepackage{orcidlink}
\usepackage[T1]{fontenc}
\usepackage[utf8]{inputenc}
\usepackage{multirow}
\usepackage{xcolor}
\usepackage{booktabs,multirow}
\usepackage{placeins}
\usepackage{threeparttable}
\usepackage{twemojis}

\usepackage{tikz}
\usetikzlibrary{patterns}
\usepackage{pgfplots}
\pgfplotsset{compat=1.18}  
\usepgfplotslibrary{fillbetween}
\usepackage{pgfplotstable}

\definecolor{authority}{HTML}{6A4C93}
\definecolor{care}{HTML}{00B4A6}
\definecolor{fairness}{HTML}{3498DB}
\definecolor{loyalty}{HTML}{E74C3C}
\definecolor{sanctity}{HTML}{F39C12}

\definecolor{twitterblue}{HTML}{1DA1F2}

\usepackage{fixme}
\fxsetup{
    status=final
}
\usepackage[inkscapeformat=pdf]{svg}
\usepackage[hyphenbreaks]{breakurl}
\usepackage[nameinlink]{cleveref}

\crefname{subfigure}{Figure}{Figures}
\Crefname{subfigure}{Figure}{Figures}

\begin{document}

\title{The Moral Gap of Large Language Models}

\author{Maciej Skórski\orcidlink{0000-0003-2997-7539}}
\affiliation{
  \institution{}
  \country{}
}
\orcid{0000-0003-2997-7539}
\email{maciej.skorski@gmail.com}

\author{Alina Landowska\orcidlink{0000-0002-7966-8243}}
\affiliation{
  \institution{}
  \country{}
}
\orcid{0000-0002-7966-8243}
\email{alina.landowska@gmail.com}

\begin{abstract}
Moral foundation detection is crucial for analyzing social discourse and developing ethically-aligned AI systems. While large language models excel across diverse tasks, their performance on specialized moral reasoning remains unclear. 

This study provides the first comprehensive comparison between state-of-the-art LLMs and fine-tuned transformers across Twitter and Reddit datasets using ROC, PR, and DET curve analysis. 

Results reveal substantial performance gaps, with LLMs exhibiting high false negative rates and systematic under-detection of moral content despite prompt engineering efforts. These findings demonstrate that task-specific fine-tuning remains superior to prompting for moral reasoning applications.

\end{abstract}

\maketitle

\section{Introduction}

\subsection{Motivation and Background}

Moral Foundations Theory (MFT) provides a framework for understanding the psychological basis of moral reasoning and judgment across cultures \cite{haidt2004intuitive}. 
Developed to explain variations in moral intuitions both within and between societies, MFT proposes that moral judgments are shaped by a set of evolved moral foundations—such as care, fairness, loyalty, authority, and sanctity—which reflect underlying psychological mechanisms \cite{haidt2012righteous,graham2013moral}.

MFT has found numerous aplications, including analysis of political ideology \cite{graham2009liberals}, environmental attitudes \cite{feinberg2013moral}, vaccine hesitancy \cite{amin2017association}, social norms \cite{forbes2020social}, news framing \cite{mokhberian2020moral}, social media discourse \cite{hoover2020moral}, everyday moral dilemmas \cite{nguyen2022mapping}, and argument assessment \cite{kobbe2020exploring, landowska2024quantitative}. 

\begin{table}[h]
\centering
\resizebox{0.99\columnwidth}{!}{
\begin{tabular}{p{0.65\columnwidth} p{0.30\columnwidth}}
\toprule
\textbf{Social Media Post} & \textbf{Moral Foundation} \\
\midrule
"My heart breaks seeing children separated from families at the border" & \textcolor{care}{\textbf{Care}} \\
"Everyone deserves equal access to healthcare regardless of income" & \textcolor{fairness}{\textbf{Fairness}} \\
"Respect your elders and follow traditional values that built this nation" & \textcolor{authority}{\textbf{Authority}} \\
"Stand with our troops - they sacrifice everything for our freedom" & \textcolor{loyalty}{\textbf{Loyalty}} \\
"Marriage is sacred and should be protected from secular corruption" & \textcolor{sanctity}{\textbf{Sanctity}} \\
\bottomrule
\end{tabular}
}
\caption{Posts and Associated Moral Foundation}
\end{table}

Computational methods for detecting moral foundations have evolved from dictionary-based approaches \cite{graham2009liberals, pennebaker1999linguistic} to transformer models \cite{preniqiMoralBERTFineTunedLanguage2024,nguyenMeasuringMoralDimensions2024a} and recently to large language models \cite{bullaLargeLanguageModels2025}. While LLMs from Anthropic and OpenAI offer appealing accessibility, no study has rigorously compared their performance to specialized transformer models on moral foundation detection. This gap leaves researchers without evidence-based guidance for tool selection, making systematic evaluation essential for informed deployment in morally sensitive applications. This paper addresses this need through comprehensive empirical comparison across multiple datasets and evaluation metrics.

\subsection{Contributions}

The key contributions to computational moral psychology are:
\begin{enumerate}
\item \textbf{Comprehensive Evaluation}: First systematic comparison of state-of-the-art LLMs (Claude Sonnet 4, GPT-o1-mini) against fine-tuned transformers (DeBERTa, RoBERTa) for moral foundation detection across Twitter and Reddit datasets, establishing performance benchmarks for both in-domain and cross-domain scenarios.
\item \textbf{Rigorous Methodology}: Establishes robust evaluation framework using the spectrum of diagnosis curves (ROC, PR, and DET) to provide comprehensive performance insights and address class-imbalance limitations of prior work that relied solely on ROC analysis.
\item \textbf{Error Analysis and Practical Guidance}: Identifies systematic LLM limitations including high false negative rates (58-90\%), foundation-specific failure patterns, and conservative prediction bias, while providing evidence-based recommendations for model selection, prompt engineering limitations, and deployment considerations in moral content analysis.
\end{enumerate}

\subsection{Related Work}

Early computational approaches to moral foundation detection relied primarily on dictionary-based methods using manually crafted lexicons such as the Moral Foundations Dictionary \citep{graham2009liberals} and its extensions \citep{Frimer_2019, hopp_extended_2021}. While still very popular due to interpretability and computationally efficiency, these approaches suffer from very low accuracy.

The advent of deep learning and particularly transformer architectures marked a significant advancement in moral content analysis. \citet{hoover2020moral} first applied deep learning models to moral foundation classification, demonstrating substantial improvements over lexicon-based approaches. Researchers have further improved accuracy by fine-tunning transformer-based models such as BERT and RoBERTa~\citep{tragerMoralFoundationsReddit2022, preniqi_moralbert_2024, nguyenMeasuringMoralDimensions2024a}, which currently achieve state-of-the-art performance.

The recent proposal of applying LLMs to moral content categorization~\cite{bullaLargeLanguageModels2025} showed promise but suffered from methodological limitations. The study used only one dataset, removed ambiguous (harder) instances, and handled annotator disagreement differently than standard practice in the field, resulting in a biased evaluation.

\newpage 

\section{Data and Methods}

\subsection{Datasets}
\begin{figure}[h]
\centering
\resizebox{\columnwidth}{!}{%
\begin{tikzpicture}
\begin{axis}[
    ybar,
    ylabel={Prevalence (\%)},
    symbolic x coords={Authority,Care,Fairness,Loyalty,Sanctity},
    xtick=data,
    legend pos=north east,
    x tick label style={rotate=45,anchor=east},
    ymin=0,ymax=45,
    nodes near coords,
    nodes near coords style={font=\small},
    width=12cm,
    height=8cm,
    axis lines*=left
]
\addplot[fill=twitterblue!80, draw=twitterblue!70] coordinates {(Authority,33.5) (Care,40.7) (Fairness,35.6) (Loyalty,30.6) (Sanctity,22.4)};
\addplot[fill=orange!70!red, draw=orange!90!red] coordinates {(Authority,19.2) (Care,26.5) (Fairness,29.5) (Loyalty,11.1) (Sanctity,9.8)};
\legend{MFTC \includegraphics[height=1.2em]{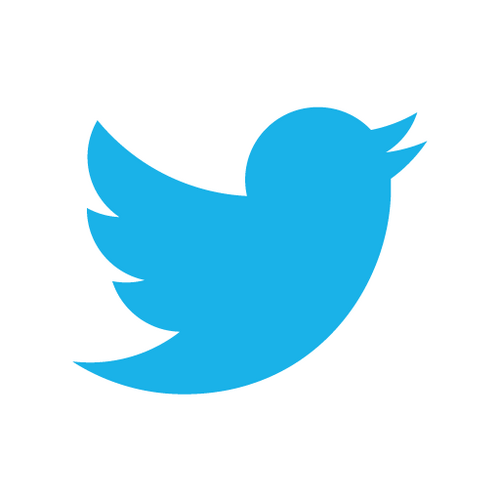},MFRC \includegraphics[height=1.2em]{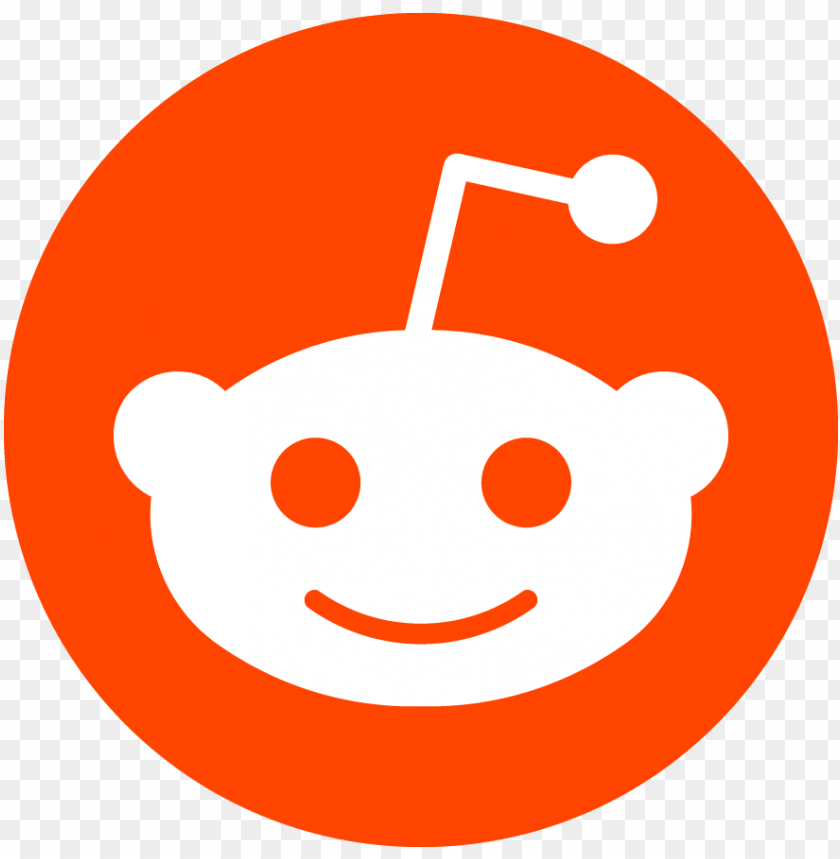}}
\end{axis}
\end{tikzpicture}
}
\caption{Moral Foundation Prevalence in Datasets}
\label{fig:prevalence_comparison}
\end{figure}

We evaluate model performance on two established moral foundations datasets. The \textbf{Twitter dataset (MFTC)} \cite{hoover2020moral} contains 34,987 tweets spanning seven socially relevant topics, with trained annotators labeling moral foundations and their sentiments. We merge virtue/vice labels (positive/negative aspects of each foundation, e.g., ``purity'' + ``degradation'' $\rightarrow$ ``sanctity''). The \textbf{Reddit dataset (MFRC)} \cite{tragerMoralFoundationsReddit2022} comprises 17,886 comments from 12 subreddits covering US politics, French politics, and everyday moral life. We merge the original equality/proportionality split back into fairness and treat ``thin morality'' cases as no foundation present. Both datasets use binary labels for five moral foundations (authority, care, fairness, loyalty, sanctity) with inclusive annotation scheme (positive if any annotator agrees). The data is summarized in \Cref{tab:datasets_percentages} and \Cref{fig:prevalence_comparison}.

\begin{table}[h]
\centering
\caption{Summary of datasets}
\label{tab:datasets_percentages}
\begin{tabular}{l l r r}
\hline
\textbf{Dataset} & \textbf{Category/Topic} & \textbf{Count} & \textbf{\%} \\
\hline
\multicolumn{4}{l}{\textbf{MFTC (Twitter)}} \\
\hline
& All Lives Matter & 4,988 & 14.3\% \\
& Black Lives Matter & 4,990 & 14.3\% \\
& 2016 US Presidential Election & 4,987 & 14.3\% \\
& Hate Speech & 4,989 & 14.3\% \\
& Hurricane Sandy & 4,990 & 14.3\% \\
& \#MeToo & 4,995 & 14.3\% \\
& Baltimore Protests & 4,985 & 14.2\% \\
\cline{2-4}
& \textbf{MFTC Subtotal} & \textbf{34,924} & \textbf{100\%} \\
\hline
\multicolumn{4}{l}{\textbf{MFRC (Reddit)}} \\
\hline
& US Politics & 6,968 & 39.0\% \\
& French Politics & 3,984 & 22.3\% \\
& Everyday Moral Life & 6,933 & 38.7\% \\
\cline{2-4}
& \textbf{MFRC Subtotal} & \textbf{17,885} & \textbf{100\%} \\
\hline
\multicolumn{3}{l}{\textbf{TOTAL ALL DATASETS}} & \textbf{52,809} \\
\hline
\end{tabular}
\end{table}

\subsection{Models}
\paragraph{Large Language Models}
We selected four recent models based on cost-effectiveness and reasoning capabilities (\Cref{tab:llm_specs}). These include general-purpose models (Haiku, GPT-4o-mini) and reasoning-specialized models (Sonnet, GPT-o1-mini) from major vendors. Models represent state-of-the-art releases: GPT-o1-mini (April 2025), Claude Sonnet (version 3.5 from February 2025 and version 4 from May 2025). All models were accessed via Python APIs with consistent prompting strategies.

\begin{table}[h]
\centering
\caption{LLM Model Specifications}
\label{tab:llm_specs}
\resizebox{\columnwidth}{!}{
\begin{tabular}{l l r r r}
\hline
\textbf{Model} & \textbf{Vendor} & \textbf{Context} & \textbf{Output} & \textbf{Price (\$/1M)} \\
\hline
Claude 3.5 Haiku & Anthropic & 200K & 4K & \$0.25 \\
Claude Sonnet 4 & Anthropic & 200K & 8K & \$3.00 \\
GPT-4o-mini & OpenAI & 128K & 16K & \$0.15 \\
GPT-o4-mini & OpenAI & 128K & 65K & \$3.00 \\
\hline
\end{tabular}
}
\footnotesize{Context and output in tokens. Pricing for input tokens (May 2025).}
\end{table}

\paragraph{Transformer Models}
While prior work established BERT models, particularly RoBERTa, as state-of-the-art for moral foundations detection, we evaluated more modern transformer architectures. For results reported in this paper, we employed DeBERTa-v3-base by Microsoft, trained with learning rate 2e-5 and the first two layers frozen for 3 epochs (throughout all results, "BERT" refers to this DeBERTa model for brevity). For in-domain evaluation, we used a 4:1 train-test split, while for out-of-domain evaluation, we trained on the full training set and evaluated on complete test sets.

\subsection{Techniques}

\paragraph{Prediction Metrics}

A range of metrics for both continuous probability scores and binary predictions is used to comprehensively evaluate model performance.
The abbreviations $TP$, $FP$, $TN$, and $FN$ represent true positives, false positives, true negatives, and false negatives, leading to:
\begin{align*}
TPR &= \frac{TP}{TP + FN}, \quad FPR = \frac{FP}{FP + TN} \\
Precision &= \frac{TP}{TP + FP}, \quad Recall = \frac{TP}{TP + FN}
\end{align*}
and aggregated metrics
\begin{align*}
F1 &= 2 \cdot \frac{Precision \times Recall}{Precision + Recall},\quad 
BER = \frac{1}{2}\left(\frac{FN}{TP + FN} + \frac{FP}{FP + TN}\right)
\end{align*}
Continuous scores are evaluated at various thresholds representing different performance tradeoffs through three complementary curves:
\begin{itemize}
\item \textbf{ROC curves}: Plot TPR vs. FPR with AUC ranging from 0.5 (random) to 1.0 (perfect)~\cite{fawcett_introduction_2006}
\item \textbf{PR curves}: Plot Precision vs. Recall with AUC considered more informative for imbalanced datasets~\cite{davis_relationship_2006}
\item \textbf{DET curves}: Plot (1-TPR) vs. FPR on normal deviate scale, emphasizing error rates~\cite{martinDETCurveAssessment1997}
\end{itemize}

\begin{figure}[h!]
\begin{tikzpicture}
\def\spacing{0.8}

\foreach \x/\colorname/\label in {
   0/care/Care,
   \spacing/fairness/Fairness,
   2*\spacing/loyalty/Loyalty,
   3*\spacing/authority/Authority,
   4*\spacing/sanctity/Sanctity
} {
  \fill[\colorname] (\x+0.6,0) -- (\x,0.15) -- (\x+0.1,0) -- (\x,-0.15) -- cycle;
  \node[above] at (\x+0.3,0.25) {\tiny\label};
}

\node[below] at (0,-0.4) {Individualism};
\node[below] at (4*\spacing+0.6,-0.4) {Collectivism};
\end{tikzpicture}
\caption{Novel moral foundations colour palette}
\label{fig:moral-palette}
\end{figure}
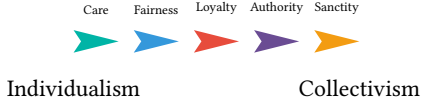

\paragraph{Moral Visualization Palette}
We have developed a novel colorblind-accessible color scheme that accurately reflects the theoretical structure of moral foundations as shown in \Cref{fig:moral-palette}. The palette represents the gradation from individualistic foundations (cooler colours) to collectivistic foundations (warmer colours): green for care as an inclusive, nurturing foundation; blue for fairness reflecting associations with justice and divine judgment; red for loyalty symbolizing bonds and allegiance; purple for authority representing traditional power and hierarchy; and gold for sanctity denoting the sacred and pure.

\section{Results}

\subsection{Detection of Moral Content}

As demonstrated in prior work, detecting whether any moral content is present, without identifying its dimensions, is relatively straightforward using text-embedding models. We reproduced (and slightly improved) results previously achieved with transformers. However, evaluation of LLMs reveals a substantial performance gap even on this simpler task. As shown in the performance curves, LLM performance is consistently inferior—they lie inside the transformer curves, indicating systematic underperformance rather than threshold-dependent differences. This evidences fundamental limitations in LLMs' ability to detect moral content compared to fine-tuned transformers.

See \Cref{fig:content-ROC} for ROC analysis and \Cref{fig:content-PR} for PR curves. The numerical metrics are reported as an additional "any" dimension when evaluating foundation-specific performance in the next section (\Cref{tab:all_results_MFRC,tab:all_results_MFTC}).

\begin{figure}[h!]
\begin{subfigure}{0.49\columnwidth}
\centering
\resizebox{\columnwidth}{!}{%
\begin{tikzpicture}
\begin{axis}[
   xlabel={False Positive Rate},
   ylabel={True Positive Rate},
   xmin=0, xmax=1,
   ymin=0, ymax=1,
   grid=major,
   grid style={gray!30},
   width=6cm,
   height=5cm,
   legend pos=south east,
   legend cell align={left},
   legend style={font=\small, nodes={scale=0.5, transform shape}, legend image post style={scale=0.7}, fill opacity=0.8,},
   axis lines=box,
    axis line style={gray!30}
]
\addplot[gray, thick] table[x=fpr, y=tpr, col sep=comma] {data/detection_in_MFRC_ROC.csv};
\addplot[mark=triangle, only marks] coordinates {(0.469,1-0.180)}; 
\addplot[mark=triangle*, only marks] coordinates {(0.369,1-0.178)}; 
\addplot[mark=square, only marks] coordinates {(0.243,1-0.352)}; 
\addplot[mark=square*, only marks] coordinates {(0.195,1-0.412)}; 
\addplot[dashed, gray] coordinates {(0,0) (1,1)};
\legend{BERT (AUC=0.86), Haiku, Sonnet, 4o-mini, o4-mini, Chance Level}
\end{axis}
\end{tikzpicture}
}
\subcaption{MFRC Dataset}
\label{fig:content-ROC-MFRC}
\end{subfigure}
\hfill
\begin{subfigure}{0.49\columnwidth}
\centering
\resizebox{\columnwidth}{!}{%
\begin{tikzpicture}
\begin{axis}[
   xlabel={False Positive Rate},
   ylabel={True Positive Rate},
   xmin=0, xmax=1,
   ymin=0, ymax=1,
   grid=major,
   grid style={gray!30},
   width=6cm,
   height=5cm,
   legend pos=south east,
   legend cell align={left},
   legend style={nodes={scale=0.5, transform shape}, legend image post style={scale=0.7}, fill opacity=0.8,},
   axis lines=box,
    axis line style={gray!30}
]
\addplot[gray, thick] table[x=fpr, y=tpr, col sep=comma] {data/detection_in_MFTC_ROC.csv};
\addplot[mark=triangle, only marks] coordinates {(0.372,1-0.158)}; 
\addplot[mark=triangle*, only marks] coordinates {(0.328,1-0.152)}; 
\addplot[mark=square, only marks] coordinates {(0.291,1-0.287)}; 
\addplot[mark=square*, only marks] coordinates {(0.204,1-0.305)}; 
\addplot[dashed, gray] coordinates {(0,0) (1,1)};
\legend{BERT (AUC=0.94), Haiku, Sonnet, gpt-4o-mini, o4-mini, Chance Level}
\end{axis}
\end{tikzpicture}
}
\subcaption{MFTC Dataset}
\label{fig:content-ROC-MFTC}
\end{subfigure}
\caption{ROC Prediction Performance.}
\label{fig:content-ROC}
\end{figure}

\begin{figure}[h!]
\begin{subfigure}{0.49\columnwidth}
\centering
\resizebox{\columnwidth}{!}{%
\begin{tikzpicture}
\begin{axis}[
   xlabel={Recall},
   ylabel={Precision},
   xmin=0, xmax=1,
   ymin=0.5, ymax=1,
   grid=major,
   grid style={gray!30},
   width=6cm,
   height=5cm,
   legend pos=south west,
   legend cell align={left},
   legend style={nodes={scale=0.5, transform shape}, legend image post style={scale=0.7}, fill opacity=0.8,},
   axis lines=box,
    axis line style={gray!30}
]
\addplot[gray, thick] table[x=recall, y=precision, col sep=comma] {data/detection_in_MFRC_PR.csv};
\addplot[mark=triangle, only marks] coordinates {(0.820, 0.696)}; 
\addplot[mark=triangle*, only marks] coordinates {(0.822, 0.745)}; 
\addplot[mark=square, only marks] coordinates {(0.648, 0.778)}; 
\addplot[mark=square*, only marks] coordinates {(0.588, 0.798)}; 
\addplot[dashed, gray] coordinates {(0,0.56) (1,0.56)};
\legend{BERT (AP=0.89), Haiku, Sonnet, 4o-mini, o4-mini, Chance Level}
\end{axis}
\end{tikzpicture}
}
\subcaption{MFRC Dataset}
\label{fig:content-PR-MFRC}
\end{subfigure}
\hfill
\begin{subfigure}{0.48\columnwidth}
\centering
\resizebox{\columnwidth}{!}{%
\begin{tikzpicture}
\begin{axis}[
   xlabel={Recall},
   ylabel={Precision},
   xmin=0, xmax=1,
   ymin=0.75, ymax=1,
   grid=major,
   grid style={gray!30},
   width=6cm,
   height=5cm,
   legend pos=south west,
   legend cell align={left},
   legend style={nodes={scale=0.5, transform shape}, legend image post style={scale=0.7}, fill opacity=0.8,},
   axis lines=box,
    axis line style={gray!30}
]
\addplot[gray, thick] table[x=recall, y=precision, col sep=comma] {data/detection_in_MFTC_PR.csv};
\addplot[mark=triangle, only marks] coordinates {(0.842, 0.896)}; 
\addplot[mark=triangle*, only marks] coordinates {(0.848, 0.908)}; 
\addplot[mark=square, only marks] coordinates {(0.713, 0.903)}; 
\addplot[mark=square*, only marks] coordinates {(0.695, 0.929)}; 
\addplot[dashed, gray] coordinates {(0,0.79) (1,0.79)};
\legend{BERT (AP=0.98), Haiku, Sonnet, gpt-4o-mini, o4-mini, Chance Level}
\end{axis}
\end{tikzpicture}
}
\subcaption{MFTC Dataset}
\label{fig:content-PR-MFTC}
\end{subfigure}
\caption{PR Prediction Performance}
\label{fig:content-PR}
\end{figure}

\subsection{Classification of Moral Values}
\input{classification}

\subsection{Impact of Prompt Engineering}

\Cref{tab:all_results_advanced_MFRC} and \Cref{tab:all_results_advanced_MFTC} show that prompt engineering generally yields improvements, with F1 gains up to +0.078 across foundations. BERT maintains substantial advantages of 0.10-0.26 F1 points even after enhanced prompting. The performance of some models in certain foundations deteriorate under prompt engineering, indicating that advanced prompting cannot reliably overcome the fundamental limitations of LLMs in moral reasoning (eee \Cref{tab:delta_results_combined}).

\begin{table}[h]
\centering
\caption{Performance on MFRC under prompt engineering}
\label{tab:all_results_advanced_MFRC}
\resizebox{0.8\columnwidth}{!}{

\begin{tabular}{lllrrrrrrr}
\toprule
 &  &  & FPR & FNR & precision & recall & f1 & ROC-AUC & AP \\
dataset & moral dimension & model &  &  &  &  &  &  &  \\
\midrule
\multirow[t]{42}{*}{MFRC} & \multirow[t]{7}{*}{any} & BERT & 0.225 & 0.221 & 0.761 & 0.862 & 0.808 & 0.862 & 0.893 \\
 &  & BERT-OOD & 0.247 & 0.306 & 0.691 & 0.892 & 0.779 & 0.800 & 0.829 \\
 &  & Haiku 3.5 & 0.622 & 0.102 & 0.655 & 0.898 & 0.757 &  &  \\
 &  & Sonnet 3.7 & 0.793 & 0.017 & 0.619 & 0.983 & 0.760 &  &  \\
 &  & Sonnet 4 & 0.626 & 0.067 & 0.662 & 0.933 & 0.774 &  &  \\
 &  & gpt-4o-mini & 0.339 & 0.310 & 0.728 & 0.690 & 0.708 &  &  \\
 &  & o4-mini & 0.622 & 0.102 & 0.655 & 0.898 & 0.757 &  &  \\
\cline{2-10}
 & \multirow[t]{7}{*}{authority} & BERT & 0.276 & 0.137 & 0.549 & 0.638 & 0.590 & 0.869 & 0.605 \\
 &  & BERT-OOD & 0.280 & 0.297 & 0.422 & 0.600 & 0.495 & 0.775 & 0.453 \\
 &  & Haiku 3.5 & 0.195 & 0.565 & 0.349 & 0.435 & 0.387 &  &  \\
 &  & Sonnet 3.7 & 0.378 & 0.245 & 0.324 & 0.755 & 0.453 &  &  \\
 &  & Sonnet 4 & 0.222 & 0.453 & 0.372 & 0.547 & 0.443 &  &  \\
 &  & gpt-4o-mini & 0.096 & 0.763 & 0.372 & 0.237 & 0.290 &  &  \\
 &  & o4-mini & 0.195 & 0.565 & 0.349 & 0.435 & 0.387 &  &  \\
\cline{2-10}
 & \multirow[t]{7}{*}{care} & BERT & 0.217 & 0.156 & 0.691 & 0.720 & 0.706 & 0.896 & 0.775 \\
 &  & BERT-OOD & 0.265 & 0.234 & 0.533 & 0.727 & 0.615 & 0.823 & 0.626 \\
 &  & Haiku 3.5 & 0.225 & 0.364 & 0.499 & 0.636 & 0.559 &  &  \\
 &  & Sonnet 3.7 & 0.291 & 0.210 & 0.488 & 0.790 & 0.603 &  &  \\
 &  & Sonnet 4 & 0.244 & 0.256 & 0.517 & 0.744 & 0.610 &  &  \\
 &  & gpt-4o-mini & 0.089 & 0.500 & 0.665 & 0.500 & 0.571 &  &  \\
 &  & o4-mini & 0.225 & 0.364 & 0.499 & 0.636 & 0.559 &  &  \\
\cline{2-10}
 & \multirow[t]{7}{*}{fairness} & BERT & 0.275 & 0.215 & 0.544 & 0.785 & 0.643 & 0.830 & 0.711 \\
 &  & BERT-OOD & 0.368 & 0.266 & 0.455 & 0.734 & 0.562 & 0.745 & 0.533 \\
 &  & Haiku 3.5 & 0.440 & 0.269 & 0.426 & 0.731 & 0.539 &  &  \\
 &  & Sonnet 3.7 & 0.434 & 0.220 & 0.446 & 0.780 & 0.567 &  &  \\
 &  & Sonnet 4 & 0.255 & 0.429 & 0.500 & 0.571 & 0.533 &  &  \\
 &  & gpt-4o-mini & 0.160 & 0.627 & 0.510 & 0.373 & 0.431 &  &  \\
 &  & o4-mini & 0.440 & 0.269 & 0.426 & 0.731 & 0.539 &  &  \\
\cline{2-10}
 & \multirow[t]{7}{*}{loyalty} & BERT & 0.194 & 0.202 & 0.471 & 0.639 & 0.542 & 0.878 & 0.563 \\
 &  & BERT-OOD & 0.243 & 0.357 & 0.372 & 0.415 & 0.392 & 0.777 & 0.395 \\
 &  & Haiku 3.5 & 0.187 & 0.607 & 0.199 & 0.393 & 0.265 &  &  \\
 &  & Sonnet 3.7 & 0.348 & 0.256 & 0.202 & 0.744 & 0.318 &  &  \\
 &  & Sonnet 4 & 0.183 & 0.446 & 0.264 & 0.554 & 0.357 &  &  \\
 &  & gpt-4o-mini & 0.066 & 0.786 & 0.277 & 0.214 & 0.241 &  &  \\
 &  & o4-mini & 0.187 & 0.607 & 0.199 & 0.393 & 0.265 &  &  \\
\cline{2-10}
 & \multirow[t]{7}{*}{sanctity} & BERT & 0.293 & 0.131 & 0.358 & 0.566 & 0.439 & 0.859 & 0.444 \\
 &  & BERT-OOD & 0.283 & 0.258 & 0.305 & 0.531 & 0.388 & 0.804 & 0.376 \\
 &  & Haiku 3.5 & 0.065 & 0.857 & 0.202 & 0.143 & 0.167 &  &  \\
 &  & Sonnet 3.7 & 0.200 & 0.430 & 0.247 & 0.570 & 0.345 &  &  \\
 &  & Sonnet 4 & 0.116 & 0.605 & 0.283 & 0.395 & 0.330 &  &  \\
 &  & gpt-4o-mini & 0.036 & 0.840 & 0.341 & 0.160 & 0.218 &  &  \\
 &  & o4-mini & 0.065 & 0.857 & 0.202 & 0.143 & 0.167 &  &  \\
\cline{1-10} \cline{2-10}
\bottomrule
\end{tabular}

}
\end{table}

\begin{table}[h]
\centering
\caption{Performance on MFTC under prompt engineering}
\label{tab:all_results_advanced_MFTC}
\resizebox{0.8\columnwidth}{!}{

\begin{tabular}{lllrrrrrrr}
\toprule
 &  &  & FPR & FNR & precision & recall & f1 & ROC-AUC & AP \\
dataset & moral dimension & model &  &  &  &  &  &  &  \\
\midrule
\multirow[t]{42}{*}{MFTC} & \multirow[t]{7}{*}{any} & BERT & 0.075 & 0.176 & 0.902 & 0.959 & 0.930 & 0.940 & 0.983 \\
 &  & BERT-OOD & 0.189 & 0.238 & 0.867 & 0.934 & 0.899 & 0.860 & 0.957 \\
 &  & Haiku 3.5 & 0.346 & 0.162 & 0.902 & 0.838 & 0.869 &  &  \\
 &  & Sonnet 3.7 & 0.643 & 0.042 & 0.850 & 0.958 & 0.901 &  &  \\
 &  & Sonnet 4 & 0.598 & 0.055 & 0.858 & 0.945 & 0.899 &  &  \\
 &  & gpt-4o-mini & 0.189 & 0.283 & 0.935 & 0.717 & 0.812 &  &  \\
 &  & o4-mini & 0.295 & 0.174 & 0.915 & 0.826 & 0.868 &  &  \\
\cline{2-10}
 & \multirow[t]{7}{*}{authority} & BERT & 0.204 & 0.173 & 0.743 & 0.742 & 0.742 & 0.899 & 0.830 \\
 &  & BERT-OOD & 0.205 & 0.395 & 0.567 & 0.645 & 0.603 & 0.749 & 0.645 \\
 &  & Haiku 3.5 & 0.140 & 0.620 & 0.565 & 0.380 & 0.454 &  &  \\
 &  & Sonnet 3.7 & 0.272 & 0.340 & 0.537 & 0.660 & 0.592 &  &  \\
 &  & Sonnet 4 & 0.173 & 0.520 & 0.570 & 0.480 & 0.521 &  &  \\
 &  & gpt-4o-mini & 0.055 & 0.769 & 0.666 & 0.231 & 0.343 &  &  \\
 &  & o4-mini & 0.167 & 0.528 & 0.574 & 0.472 & 0.518 &  &  \\
\cline{2-10}
 & \multirow[t]{7}{*}{care} & BERT & 0.115 & 0.244 & 0.805 & 0.769 & 0.787 & 0.897 & 0.881 \\
 &  & BERT-OOD & 0.252 & 0.261 & 0.663 & 0.748 & 0.703 & 0.811 & 0.765 \\
 &  & Haiku 3.5 & 0.334 & 0.322 & 0.586 & 0.678 & 0.629 &  &  \\
 &  & Sonnet 3.7 & 0.436 & 0.173 & 0.570 & 0.827 & 0.675 &  &  \\
 &  & Sonnet 4 & 0.376 & 0.218 & 0.592 & 0.782 & 0.674 &  &  \\
 &  & gpt-4o-mini & 0.164 & 0.473 & 0.691 & 0.527 & 0.598 &  &  \\
 &  & o4-mini & 0.243 & 0.341 & 0.654 & 0.659 & 0.656 &  &  \\
\cline{2-10}
 & \multirow[t]{7}{*}{fairness} & BERT & 0.111 & 0.232 & 0.792 & 0.768 & 0.780 & 0.906 & 0.874 \\
 &  & BERT-OOD & 0.317 & 0.286 & 0.512 & 0.819 & 0.630 & 0.770 & 0.659 \\
 &  & Haiku 3.5 & 0.289 & 0.280 & 0.580 & 0.720 & 0.642 &  &  \\
 &  & Sonnet 3.7 & 0.307 & 0.193 & 0.592 & 0.807 & 0.683 &  &  \\
 &  & Sonnet 4 & 0.232 & 0.301 & 0.624 & 0.699 & 0.659 &  &  \\
 &  & gpt-4o-mini & 0.140 & 0.518 & 0.655 & 0.482 & 0.556 &  &  \\
 &  & o4-mini & 0.268 & 0.287 & 0.595 & 0.713 & 0.649 &  &  \\
\cline{2-10}
 & \multirow[t]{7}{*}{loyalty} & BERT & 0.159 & 0.267 & 0.670 & 0.733 & 0.700 & 0.866 & 0.782 \\
 &  & BERT-OOD & 0.228 & 0.460 & 0.442 & 0.680 & 0.536 & 0.713 & 0.551 \\
 &  & Haiku 3.5 & 0.129 & 0.703 & 0.498 & 0.297 & 0.372 &  &  \\
 &  & Sonnet 3.7 & 0.268 & 0.441 & 0.473 & 0.559 & 0.513 &  &  \\
 &  & Sonnet 4 & 0.156 & 0.612 & 0.516 & 0.388 & 0.443 &  &  \\
 &  & gpt-4o-mini & 0.075 & 0.732 & 0.606 & 0.268 & 0.371 &  &  \\
 &  & o4-mini & 0.148 & 0.613 & 0.529 & 0.387 & 0.447 &  &  \\
\cline{2-10}
 & \multirow[t]{7}{*}{sanctity} & BERT & 0.138 & 0.262 & 0.688 & 0.654 & 0.671 & 0.875 & 0.740 \\
 &  & BERT-OOD & 0.340 & 0.271 & 0.433 & 0.616 & 0.509 & 0.761 & 0.523 \\
 &  & Haiku 3.5 & 0.059 & 0.731 & 0.570 & 0.269 & 0.366 &  &  \\
 &  & Sonnet 3.7 & 0.213 & 0.419 & 0.441 & 0.581 & 0.502 &  &  \\
 &  & Sonnet 4 & 0.219 & 0.417 & 0.435 & 0.583 & 0.498 &  &  \\
 &  & gpt-4o-mini & 0.030 & 0.776 & 0.687 & 0.224 & 0.338 &  &  \\
 &  & o4-mini & 0.067 & 0.663 & 0.591 & 0.337 & 0.429 &  &  \\
\cline{1-10} \cline{2-10}
\bottomrule
\end{tabular}

}
\end{table}

\begin{table}[htbp]
   \caption{Prompt engineering performance delta comparison}
   \label{tab:delta_results_combined}
   \centering
   \footnotesize
\resizebox{0.7\columnwidth}{!}{
   \begin{tabular}{@{}ll@{\hspace{10pt}}l@{\hspace{8pt}}r@{\hspace{8pt}}r@{\hspace{8pt}}r@{\hspace{8pt}}r@{\hspace{8pt}}r@{}}
   \toprule
   \textbf{Dataset} & \textbf{Moral Dimension} & \textbf{Model} & \textbf{$\Delta$ FPR} & \textbf{$\Delta$ FNR} & \textbf{$\Delta$ Prec} & \textbf{$\Delta$ Rec} & \textbf{$\Delta$ F1} \\
   \midrule
   \multirow{24}{*}{MFTC} 
   & \multirow{4}{*}{any} 
   & Haiku 3.5     & \textcolor{blue}{$-0.03$} & \textcolor{orange}{$0.00$} & \textcolor{blue}{$0.01$} & \textcolor{orange}{$0.00$} & \textcolor{blue}{$0.00$} \\
   && Sonnet 4       & \textcolor{orange}{$0.27$} & \textcolor{blue}{$-0.10$} & \textcolor{orange}{$-0.05$} & \textcolor{blue}{$0.10$} & \textcolor{blue}{$0.02$} \\
   && gpt-4o-mini    & \textcolor{blue}{$-0.10$} & \textcolor{blue}{$0.00$} & \textcolor{blue}{$0.03$} & \textcolor{blue}{$0.00$} & \textcolor{blue}{$0.02$} \\
   && o4-mini        & \textcolor{orange}{$0.09$} & \textcolor{blue}{$-0.13$} & \textcolor{orange}{$-0.01$} & \textcolor{blue}{$0.13$} & \textcolor{blue}{$0.07$} \\
   \cmidrule{2-8}
   & \multirow{4}{*}{authority} 
   & Haiku 3.5     & \textcolor{blue}{$-0.01$} & \textcolor{orange}{$0.01$} & \textcolor{blue}{$0.01$} & \textcolor{orange}{$-0.01$} & \textcolor{orange}{$0.00$} \\
   && Sonnet 4       & \textcolor{orange}{$0.03$} & \textcolor{blue}{$-0.06$} & \textcolor{orange}{$-0.02$} & \textcolor{blue}{$0.06$} & \textcolor{blue}{$0.03$} \\
   && gpt-4o-mini    & \textcolor{blue}{$-0.15$} & \textcolor{orange}{$0.17$} & \textcolor{blue}{$0.18$} & \textcolor{orange}{$-0.17$} & \textcolor{orange}{$-0.09$} \\
   && o4-mini        & \textcolor{orange}{$0.02$} & \textcolor{blue}{$-0.17$} & \textcolor{blue}{$0.08$} & \textcolor{blue}{$0.17$} & \textcolor{blue}{$0.14$} \\
   \cmidrule{2-8}
   & \multirow{4}{*}{care} 
   & Haiku 3.5     & \textcolor{orange}{$0.02$} & \textcolor{blue}{$-0.02$} & \textcolor{orange}{$-0.01$} & \textcolor{blue}{$0.02$} & \textcolor{blue}{$0.01$} \\
   && Sonnet 4       & \textcolor{orange}{$0.04$} & \textcolor{blue}{$-0.04$} & \textcolor{orange}{$-0.01$} & \textcolor{blue}{$0.04$} & \textcolor{blue}{$0.01$} \\
   && gpt-4o-mini    & \textcolor{blue}{$-0.09$} & \textcolor{orange}{$0.03$} & \textcolor{blue}{$0.09$} & \textcolor{orange}{$-0.03$} & \textcolor{blue}{$0.02$} \\
   && o4-mini        & \textcolor{orange}{$0.02$} & \textcolor{blue}{$-0.06$} & \textcolor{blue}{$0.00$} & \textcolor{blue}{$0.06$} & \textcolor{blue}{$0.03$} \\
   \cmidrule{2-8}
   & \multirow{4}{*}{fairness} 
   & Haiku 3.5     & \textcolor{orange}{$0.05$} & \textcolor{blue}{$-0.06$} & \textcolor{orange}{$-0.02$} & \textcolor{blue}{$0.06$} & \textcolor{blue}{$0.01$} \\
   && Sonnet 4       & \textcolor{orange}{$0.04$} & \textcolor{blue}{$-0.02$} & \textcolor{orange}{$-0.03$} & \textcolor{blue}{$0.02$} & \textcolor{orange}{$-0.01$} \\
   && gpt-4o-mini    & \textcolor{orange}{$0.01$} & \textcolor{blue}{$-0.22$} & \textcolor{blue}{$0.12$} & \textcolor{blue}{$0.22$} & \textcolor{blue}{$0.20$} \\
   && o4-mini        & \textcolor{orange}{$0.08$} & \textcolor{blue}{$-0.15$} & \textcolor{orange}{$-0.03$} & \textcolor{blue}{$0.15$} & \textcolor{blue}{$0.06$} \\
   \cmidrule{2-8}
   & \multirow{4}{*}{loyalty} 
   & Haiku 3.5     & \textcolor{orange}{$0.05$} & \textcolor{blue}{$-0.12$} & \textcolor{blue}{$0.01$} & \textcolor{blue}{$0.12$} & \textcolor{blue}{$0.11$} \\
   && Sonnet 4       & \textcolor{orange}{$0.07$} & \textcolor{blue}{$-0.06$} & \textcolor{orange}{$-0.11$} & \textcolor{blue}{$0.06$} & \textcolor{blue}{$0.02$} \\
   && gpt-4o-mini    & \textcolor{blue}{$-0.02$} & \textcolor{blue}{$-0.15$} & \textcolor{blue}{$0.25$} & \textcolor{blue}{$0.15$} & \textcolor{blue}{$0.19$} \\
   && o4-mini        & \textcolor{orange}{$0.02$} & \textcolor{blue}{$-0.17$} & \textcolor{blue}{$0.10$} & \textcolor{blue}{$0.17$} & \textcolor{blue}{$0.16$} \\
   \cmidrule{2-8}
   & \multirow{4}{*}{sanctity} 
   & Haiku 3.5     & \textcolor{blue}{$-0.06$} & \textcolor{blue}{$-0.03$} & \textcolor{blue}{$0.21$} & \textcolor{blue}{$0.03$} & \textcolor{blue}{$0.08$} \\
   && Sonnet 4       & \textcolor{orange}{$0.12$} & \textcolor{blue}{$-0.23$} & \textcolor{orange}{$-0.08$} & \textcolor{blue}{$0.23$} & \textcolor{blue}{$0.08$} \\
   && gpt-4o-mini    & \textcolor{blue}{$-0.07$} & \textcolor{blue}{$-0.06$} & \textcolor{blue}{$0.36$} & \textcolor{blue}{$0.06$} & \textcolor{blue}{$0.12$} \\
   && o4-mini        & \textcolor{blue}{$-0.05$} & \textcolor{blue}{$-0.12$} & \textcolor{blue}{$0.24$} & \textcolor{blue}{$0.12$} & \textcolor{blue}{$0.16$} \\
   \midrule
   \multirow{20}{*}{MFRC} 
   & \multirow{4}{*}{any} 
   & Haiku 3.5      & \textcolor{orange}{$0.15$} & \textcolor{blue}{$-0.08$} & \textcolor{orange}{$-0.04$} & \textcolor{blue}{$0.08$} & \textcolor{orange}{$0.00$} \\
   && Sonnet 4        & \textcolor{orange}{$0.27$} & \textcolor{blue}{$-0.11$} & \textcolor{orange}{$-0.09$} & \textcolor{blue}{$0.11$} & \textcolor{orange}{$-0.01$} \\
   && gpt-4o-mini     & \textcolor{orange}{$0.10$} & \textcolor{blue}{$-0.04$} & \textcolor{orange}{$-0.05$} & \textcolor{blue}{$0.04$} & \textcolor{orange}{$0.00$} \\
   && o4-mini         & \textcolor{orange}{$0.43$} & \textcolor{blue}{$-0.31$} & \textcolor{orange}{$-0.14$} & \textcolor{blue}{$0.31$} & \textcolor{blue}{$0.08$} \\
   \cmidrule{2-8}
   & \multirow{4}{*}{authority} 
   & Haiku 3.5      & \textcolor{blue}{$-0.03$} & \textcolor{orange}{$0.04$} & \textcolor{blue}{$0.02$} & \textcolor{orange}{$-0.04$} & \textcolor{orange}{$0.00$} \\
   && Sonnet 4        & \textcolor{orange}{$0.09$} & \textcolor{blue}{$-0.12$} & \textcolor{orange}{$-0.05$} & \textcolor{blue}{$0.12$} & \textcolor{blue}{$0.02$} \\
   && gpt-4o-mini     & \textcolor{blue}{$-0.05$} & \textcolor{orange}{$0.14$} & \textcolor{orange}{$-0.01$} & \textcolor{orange}{$-0.14$} & \textcolor{orange}{$-0.09$} \\
   && o4-mini         & \textcolor{orange}{$0.10$} & \textcolor{blue}{$-0.20$} & \textcolor{orange}{$-0.03$} & \textcolor{blue}{$0.20$} & \textcolor{blue}{$0.10$} \\
   \cmidrule{2-8}
   & \multirow{4}{*}{care} 
   & Haiku 3.5      & \textcolor{blue}{$-0.03$} & \textcolor{blue}{$-0.04$} & \textcolor{blue}{$0.05$} & \textcolor{blue}{$0.04$} & \textcolor{blue}{$0.05$} \\
   && Sonnet 4        & \textcolor{orange}{$0.04$} & \textcolor{blue}{$-0.05$} & \textcolor{orange}{$-0.03$} & \textcolor{blue}{$0.05$} & \textcolor{orange}{$0.00$} \\
   && gpt-4o-mini     & \textcolor{blue}{$-0.05$} & \textcolor{orange}{$0.02$} & \textcolor{blue}{$0.09$} & \textcolor{orange}{$-0.02$} & \textcolor{blue}{$0.02$} \\
   && o4-mini         & \textcolor{orange}{$0.07$} & \textcolor{blue}{$-0.11$} & \textcolor{orange}{$-0.05$} & \textcolor{blue}{$0.11$} & \textcolor{blue}{$0.02$} \\
   \cmidrule{2-8}
   & \multirow{4}{*}{fairness} 
   & Haiku 3.5      & \textcolor{orange}{$0.08$} & \textcolor{blue}{$-0.15$} & \textcolor{blue}{$0.01$} & \textcolor{blue}{$0.15$} & \textcolor{blue}{$0.05$} \\
   && Sonnet 4        & \textcolor{orange}{$0.06$} & \textcolor{blue}{$-0.09$} & \textcolor{orange}{$-0.03$} & \textcolor{blue}{$0.09$} & \textcolor{blue}{$0.03$} \\
   && gpt-4o-mini     & \textcolor{orange}{$0.03$} & \textcolor{blue}{$-0.14$} & \textcolor{blue}{$0.05$} & \textcolor{blue}{$0.14$} & \textcolor{blue}{$0.12$} \\
   && o4-mini         & \textcolor{orange}{$0.29$} & \textcolor{blue}{$-0.35$} & \textcolor{orange}{$-0.11$} & \textcolor{blue}{$0.35$} & \textcolor{blue}{$0.09$} \\
   \cmidrule{2-8}
   & \multirow{4}{*}{loyalty} 
   & Haiku 3.5      & \textcolor{orange}{$0.04$} & \textcolor{blue}{$-0.08$} & \textcolor{orange}{$0.00$} & \textcolor{blue}{$0.08$} & \textcolor{blue}{$0.02$} \\
   && Sonnet 4        & \textcolor{orange}{$0.07$} & \textcolor{blue}{$-0.07$} & \textcolor{orange}{$-0.09$} & \textcolor{blue}{$0.07$} & \textcolor{orange}{$-0.04$} \\
   && gpt-4o-mini     & \textcolor{orange}{$0.01$} & \textcolor{blue}{$-0.11$} & \textcolor{blue}{$0.08$} & \textcolor{blue}{$0.11$} & \textcolor{blue}{$0.10$} \\
   && o4-mini         & \textcolor{orange}{$0.11$} & \textcolor{blue}{$-0.19$} & \textcolor{orange}{$-0.05$} & \textcolor{blue}{$0.19$} & \textcolor{blue}{$0.04$} \\
   \cmidrule{2-8}
   & \multirow{4}{*}{sanctity} 
   & Haiku 3.5      & \textcolor{blue}{$-0.10$} & \textcolor{orange}{$0.14$} & \textcolor{blue}{$0.03$} & \textcolor{orange}{$-0.14$} & \textcolor{orange}{$-0.05$} \\
   && Sonnet 4        & \textcolor{orange}{$0.06$} & \textcolor{blue}{$-0.11$} & \textcolor{orange}{$-0.08$} & \textcolor{blue}{$0.11$} & \textcolor{orange}{$0.00$} \\
   && gpt-4o-mini     & \textcolor{blue}{$-0.02$} & \textcolor{blue}{$-0.06$} & \textcolor{blue}{$0.18$} & \textcolor{blue}{$0.06$} & \textcolor{blue}{$0.10$} \\
   && o4-mini         & \textcolor{blue}{$-0.01$} & \textcolor{orange}{$0.00$} & \textcolor{blue}{$0.02$} & \textcolor{orange}{$0.00$} & \textcolor{blue}{$0.01$} \\
   \bottomrule
   \end{tabular}
}
   \vspace{0.3cm}
   \begin{minipage}{\linewidth}
   \footnotesize
   \textbf{Notes:} $\Delta$ = Prompt-Engineered $-$ Basic Prompt. 
   \textcolor{blue}{Blue} = improvement, \textcolor{orange}{Orange} = degradation.
   For FPR/FNR: negative is improvement; for Precision/Recall/F1: positive is improvement.
   \end{minipage}
\end{table}


\begin{figure*}[h!]
\centering
\begin{subfigure}{0.49\textwidth}
\centering
\begin{tikzpicture}
\begin{axis}[
   height=6cm,
   width=8cm,
   xlabel={False Positive Rate},
   ylabel={True Positive Rate},
   xmin=0, xmax=1,
   ymin=0, ymax=1,
   grid=major,
   legend pos=south east,
   legend cell align={left},
   legend style={font=\tiny, legend image post style={scale=0.8}, fill opacity=0.8,}
]

\addlegendimage{draw, black}
\addlegendentry{BERT}
\addlegendimage{mark=triangle, only marks, black}
\addlegendentry{Haiku}
\addlegendimage{mark=triangle*, only marks, black}
\addlegendentry{Sonnet}
\addlegendimage{mark=square, only marks, black}
\addlegendentry{gpt-4o-mini}
\addlegendimage{mark=square*, only marks, black}
\addlegendentry{o4-mini}
\addlegendimage{dashed, gray}
\addlegendentry{Chance}

\addplot[care] table[x=fpr-care, y=tpr-care, col sep=comma] {data/classification_MFRC_ROC.csv};
\addplot[mark=triangle, only marks, care, thick] coordinates {(0.225,{1-0.364})}; 
\addplot[mark=triangle*, only marks, care] coordinates {(0.244,{1-0.256})}; 
\addplot[mark=square, only marks, care, thick] coordinates {(0.089,{1-0.500})}; 
\addplot[mark=square*, only marks, care] coordinates {(0.225,{1-0.364})}; 

\addplot[loyalty] table[x=fpr-loyalty, y=tpr-loyalty, col sep=comma] {data/classification_MFRC_ROC.csv};
\addplot[mark=triangle, only marks, loyalty, thick] coordinates {(0.187,{1-0.607})}; 
\addplot[mark=triangle*, only marks, loyalty] coordinates {(0.183,{1-0.446})}; 
\addplot[mark=square, only marks, loyalty, thick] coordinates {(0.066,{1-0.786})}; 
\addplot[mark=square*, only marks, loyalty] coordinates {(0.187,{1-0.607})}; 

\addplot[fairness] table[x=fpr-fairness, y=tpr-fairness, col sep=comma] {data/classification_MFRC_ROC.csv};
\addplot[mark=triangle, only marks, fairness, thick] coordinates {(0.440,{1-0.269})}; 
\addplot[mark=triangle*, only marks, fairness] coordinates {(0.255,{1-0.429})}; 
\addplot[mark=square, only marks, fairness, thick] coordinates {(0.160,{1-0.627})}; 
\addplot[mark=square*, only marks, fairness] coordinates {(0.440,{1-0.269})}; 

\addplot[authority] table[x=fpr-authority, y=tpr-authority, col sep=comma] {data/classification_MFRC_ROC.csv};
\addplot[mark=triangle, only marks, authority, thick] coordinates {(0.195,{1-0.565})}; 
\addplot[mark=triangle*, only marks, authority] coordinates {(0.222,{1-0.453})}; 
\addplot[mark=square, only marks, authority, thick] coordinates {(0.096,{1-0.763})}; 
\addplot[mark=square*, only marks, authority] coordinates {(0.195,{1-0.565})}; 

\addplot[sanctity, thick] table[x=fpr-sanctity, y=tpr-sanctity, col sep=comma] {data/classification_MFRC_ROC.csv};
\addplot[mark=triangle, only marks, sanctity, thick] coordinates {(0.065,{1-0.857})}; 
\addplot[mark=triangle*, only marks, sanctity] coordinates {(0.116,{1-0.605})}; 
\addplot[mark=square, only marks, sanctity, thick] coordinates {(0.036,{1-0.840})}; 
\addplot[mark=square*, only marks, sanctity] coordinates {(0.065,{1-0.857})}; 

\addplot[dashed, gray] coordinates {(0,0) (1,1)};

\end{axis}
\end{tikzpicture}
\begin{center}
\small
\textcolor{care}{\rule{0.4cm}{2pt}} Care \quad  
\textcolor{fairness}{\rule{0.4cm}{2pt}} Fairness \quad
\textcolor{loyalty}{\rule{0.4cm}{2pt}} Loyalty \quad
\textcolor{authority}{\rule{0.4cm}{2pt}} Authority \quad
\textcolor{sanctity}{\rule{0.4cm}{2pt}} Sanctity
\end{center}
\caption{ROC curves for MFRC dataset.}
\label{fig:ROC-MFRC-advanced}
\end{subfigure}\hfill
\begin{subfigure}{0.49\textwidth}
\centering
\begin{tikzpicture}
\begin{axis}[
   height=6cm,
   width=8cm,
   xlabel={False Positive Rate},
   ylabel={True Positive Rate},
   xmin=0, xmax=1,
   ymin=0, ymax=1,
   grid=major,
   legend pos=south east,
   legend cell align={left},
   legend style={font=\tiny, legend image post style={scale=0.8}}
]

\addlegendimage{draw, black}
\addlegendentry{BERT}
\addlegendimage{mark=triangle, only marks, black}
\addlegendentry{Haiku}
\addlegendimage{mark=triangle*, only marks, black}
\addlegendentry{Sonnet}
\addlegendimage{mark=square, only marks, black}
\addlegendentry{gpt-4o-mini}
\addlegendimage{mark=square*, only marks, black}
\addlegendentry{o4-mini}
\addlegendimage{dashed, gray}
\addlegendentry{Chance}

\addplot[care] table[x=fpr-care, y=tpr-care, col sep=comma] {data/classification_MFTC_ROC.csv};
\addplot[mark=triangle, only marks, care, thick] coordinates {(0.334,{1-0.322})}; 
\addplot[mark=triangle*, only marks, care] coordinates {(0.376,{1-0.218})}; 
\addplot[mark=square, only marks, care, thick] coordinates {(0.164,{1-0.473})}; 
\addplot[mark=square*, only marks, care] coordinates {(0.243,{1-0.341})}; 

\addplot[loyalty] table[x=fpr-loyalty, y=tpr-loyalty, col sep=comma] {data/classification_MFTC_ROC.csv};
\addplot[mark=triangle, only marks, loyalty, thick] coordinates {(0.129,{1-0.703})}; 
\addplot[mark=triangle*, only marks, loyalty] coordinates {(0.156,{1-0.612})}; 
\addplot[mark=square, only marks, loyalty, thick] coordinates {(0.075,{1-0.732})}; 
\addplot[mark=square*, only marks, loyalty] coordinates {(0.148,{1-0.613})}; 

\addplot[fairness] table[x=fpr-fairness, y=tpr-fairness, col sep=comma] {data/classification_MFTC_ROC.csv};
\addplot[mark=triangle, only marks, fairness, thick] coordinates {(0.289,{1-0.280})}; 
\addplot[mark=triangle*, only marks, fairness] coordinates {(0.232,{1-0.301})}; 
\addplot[mark=square, only marks, fairness, thick] coordinates {(0.140,{1-0.518})}; 
\addplot[mark=square*, only marks, fairness] coordinates {(0.268,{1-0.287})}; 

\addplot[authority] table[x=fpr-authority, y=tpr-authority, col sep=comma] {data/classification_MFTC_ROC.csv};
\addplot[mark=triangle, only marks, authority, thick] coordinates {(0.140,{1-0.620})}; 
\addplot[mark=triangle*, only marks, authority] coordinates {(0.173,{1-0.520})}; 
\addplot[mark=square, only marks, authority, thick] coordinates {(0.055,{1-0.769})}; 
\addplot[mark=square*, only marks, authority] coordinates {(0.167,{1-0.528})}; 

\addplot[sanctity, thick] table[x=fpr-sanctity, y=tpr-sanctity, col sep=comma] {data/classification_MFTC_ROC.csv};
\addplot[mark=triangle, only marks, sanctity, thick] coordinates {(0.059,{1-0.731})}; 
\addplot[mark=triangle*, only marks, sanctity] coordinates {(0.219,{1-0.417})}; 
\addplot[mark=square, only marks, sanctity, thick] coordinates {(0.030,{1-0.776})}; 
\addplot[mark=square*, only marks, sanctity] coordinates {(0.067,{1-0.663})}; 

\addplot[dashed, gray] coordinates {(0,0) (1,1)};

\end{axis}
\end{tikzpicture}
\begin{center}
\small
\textcolor{care}{\rule{0.4cm}{2pt}} Care \quad  
\textcolor{fairness}{\rule{0.4cm}{2pt}} Fairness \quad
\textcolor{loyalty}{\rule{0.4cm}{2pt}} Loyalty \quad
\textcolor{authority}{\rule{0.4cm}{2pt}} Authority \quad
\textcolor{sanctity}{\rule{0.4cm}{2pt}} Sanctity
\end{center}
\caption{ROC curves for MFTC Dataset}
\label{fig:ROC-MFTC-advanced}
\end{subfigure}
\label{fig:ROC-combined-advanced}
\end{figure*}


\begin{figure*}[h!]
\centering
\begin{subfigure}{0.49\textwidth}
\centering
\begin{tikzpicture}
\begin{axis}[
   height=6cm,
   width=8cm,
   xlabel={Recall},
   ylabel={Precision},
   xmin=0, xmax=1,
   ymin=0, ymax=1,
   grid=major,
   legend pos=north east,
   legend cell align={left},
   legend style={font=\tiny, legend image post style={scale=0.8}, fill opacity=0.6,}
]
\addlegendimage{draw, black}
\addlegendentry{BERT}
\addlegendimage{mark=triangle, only marks, black}
\addlegendentry{Haiku}
\addlegendimage{mark=triangle*, only marks, black}
\addlegendentry{Sonnet}
\addlegendimage{mark=square, only marks, black}
\addlegendentry{gpt-4o-mini}
\addlegendimage{mark=square*, only marks, black}
\addlegendentry{o4-mini}

\addplot[care] table[x=recall-care, y=precision-care, col sep=comma] {data/classification_MFRC_PR.csv};
\addplot[mark=triangle, only marks, care, thick] coordinates {(0.636,0.499)}; 
\addplot[mark=triangle*, only marks, care] coordinates {(0.744,0.517)}; 
\addplot[mark=square, only marks, care, thick] coordinates {(0.500,0.665)}; 
\addplot[mark=square*, only marks, care] coordinates {(0.636,0.499)}; 

\addplot[loyalty] table[x=recall-loyalty, y=precision-loyalty, col sep=comma] {data/classification_MFRC_PR.csv};
\addplot[mark=triangle, only marks, loyalty, thick] coordinates {(0.393,0.199)}; 
\addplot[mark=triangle*, only marks, loyalty] coordinates {(0.554,0.264)}; 
\addplot[mark=square, only marks, loyalty, thick] coordinates {(0.214,0.277)}; 
\addplot[mark=square*, only marks, loyalty] coordinates {(0.393,0.199)}; 

\addplot[fairness] table[x=recall-fairness, y=precision-fairness, col sep=comma] {data/classification_MFRC_PR.csv};
\addplot[mark=triangle, only marks, fairness, thick] coordinates {(0.731,0.426)}; 
\addplot[mark=triangle*, only marks, fairness] coordinates {(0.571,0.500)}; 
\addplot[mark=square, only marks, fairness, thick] coordinates {(0.373,0.510)}; 
\addplot[mark=square*, only marks, fairness] coordinates {(0.731,0.426)}; 

\addplot[authority] table[x=recall-authority, y=precision-authority, col sep=comma] {data/classification_MFRC_PR.csv};
\addplot[mark=triangle, only marks, authority, thick] coordinates {(0.435,0.349)}; 
\addplot[mark=triangle*, only marks, authority] coordinates {(0.547,0.372)}; 
\addplot[mark=square, only marks, authority, thick] coordinates {(0.237,0.372)}; 
\addplot[mark=square*, only marks, authority] coordinates {(0.435,0.349)}; 

\addplot[sanctity, thick] table[x=recall-sanctity, y=precision-sanctity, col sep=comma] {data/classification_MFRC_PR.csv};
\addplot[mark=triangle, only marks, sanctity, thick] coordinates {(0.143,0.202)}; 
\addplot[mark=triangle*, only marks, sanctity] coordinates {(0.395,0.283)}; 
\addplot[mark=square, only marks, sanctity, thick] coordinates {(0.160,0.341)}; 
\addplot[mark=square*, only marks, sanctity] coordinates {(0.143,0.202)}; 

\end{axis}
\end{tikzpicture}
\begin{center}
\small
\textcolor{care}{\rule{0.4cm}{2pt}} Care \quad  
\textcolor{fairness}{\rule{0.4cm}{2pt}} Fairness \quad
\textcolor{loyalty}{\rule{0.4cm}{2pt}} Loyalty \quad
\textcolor{authority}{\rule{0.4cm}{2pt}} Authority \quad
\textcolor{sanctity}{\rule{0.4cm}{2pt}} Sanctity
\end{center}
\caption{Precision-recall curves for MFRC dataset}
\label{fig:PR-MFRC-advanced}
\end{subfigure}\hfill
\begin{subfigure}{0.49\textwidth}
\centering
\begin{tikzpicture}
\begin{axis}[
   height=6cm,
   width=8cm,
   xlabel={Recall},
   ylabel={Precision},
   xmin=0, xmax=1,
   ymin=0, ymax=1,
   grid=major,
   legend pos=north east,
   legend cell align={left},
   legend style={font=\tiny, legend image post style={scale=0.8}, fill opacity=0.8,}
]
\addlegendimage{draw, black}
\addlegendentry{BERT}
\addlegendimage{mark=triangle, only marks, black}
\addlegendentry{Haiku}
\addlegendimage{mark=triangle*, only marks, black}
\addlegendentry{Sonnet}
\addlegendimage{mark=square, only marks, black}
\addlegendentry{gpt-4o-mini}
\addlegendimage{mark=square*, only marks, black}
\addlegendentry{o4-mini}

\addplot[care] table[x=recall-care, y=precision-care, col sep=comma] {data/classification_MFTC_PR.csv};
\addplot[mark=triangle, only marks, care, thick] coordinates {(0.678,0.586)}; 
\addplot[mark=triangle*, only marks, care] coordinates {(0.782,0.592)}; 
\addplot[mark=square, only marks, care, thick] coordinates {(0.527,0.691)}; 
\addplot[mark=square*, only marks, care] coordinates {(0.659,0.654)}; 

\addplot[loyalty] table[x=recall-loyalty, y=precision-loyalty, col sep=comma] {data/classification_MFTC_PR.csv};
\addplot[mark=triangle, only marks, loyalty, thick] coordinates {(0.297,0.498)}; 
\addplot[mark=triangle*, only marks, loyalty] coordinates {(0.388,0.516)}; 
\addplot[mark=square, only marks, loyalty, thick] coordinates {(0.268,0.606)}; 
\addplot[mark=square*, only marks, loyalty] coordinates {(0.387,0.529)}; 

\addplot[fairness] table[x=recall-fairness, y=precision-fairness, col sep=comma] {data/classification_MFTC_PR.csv};
\addplot[mark=triangle, only marks, fairness, thick] coordinates {(0.720,0.580)}; 
\addplot[mark=triangle*, only marks, fairness] coordinates {(0.699,0.624)}; 
\addplot[mark=square, only marks, fairness, thick] coordinates {(0.482,0.655)}; 
\addplot[mark=square*, only marks, fairness] coordinates {(0.713,0.595)}; 

\addplot[authority] table[x=recall-authority, y=precision-authority, col sep=comma] {data/classification_MFTC_PR.csv};
\addplot[mark=triangle, only marks, authority, thick] coordinates {(0.380,0.565)}; 
\addplot[mark=triangle*, only marks, authority] coordinates {(0.480,0.570)}; 
\addplot[mark=square, only marks, authority, thick] coordinates {(0.231,0.666)}; 
\addplot[mark=square*, only marks, authority] coordinates {(0.472,0.574)}; 

\addplot[sanctity, thick] table[x=recall-sanctity, y=precision-sanctity, col sep=comma] {data/classification_MFTC_PR.csv};
\addplot[mark=triangle, only marks, sanctity, thick] coordinates {(0.269,0.570)}; 
\addplot[mark=triangle*, only marks, sanctity] coordinates {(0.583,0.435)}; 
\addplot[mark=square, only marks, sanctity, thick] coordinates {(0.224,0.687)}; 
\addplot[mark=square*, only marks, sanctity] coordinates {(0.337,0.591)}; 

\end{axis}
\end{tikzpicture}
\begin{center}
\small
\textcolor{care}{\rule{0.4cm}{2pt}} Care \quad  
\textcolor{fairness}{\rule{0.4cm}{2pt}} Fairness \quad
\textcolor{loyalty}{\rule{0.4cm}{2pt}} Loyalty \quad
\textcolor{authority}{\rule{0.4cm}{2pt}} Authority \quad
\textcolor{sanctity}{\rule{0.4cm}{2pt}} Sanctity
\end{center}
\caption{Precision-recall curves for MFTC dataset}
\label{fig:PR-MFTC-advanced}
\end{subfigure}
\end{figure*}

\begin{figure*}[h!]
\centering
\begin{subfigure}{0.49\textwidth}
\centering
\begin{tikzpicture}
\begin{axis}[
   height=6cm,
   width=8cm,
    xlabel={False Positive Rate (Normal Deviate)},
    ylabel={False Negative Rate (Normal Deviate)},
    xmin=-2.33, xmax=1.64,
    ymin=-2.33, ymax=1.64,
    grid=major,
    legend pos=north east,
    legend cell align={left},
    legend style={font=\tiny, legend image post style={scale=0.8}, fill opacity=0.8},
    xtick={-2.33, -1.64, -1.28, -0.84, 0, 0.84, 1.28, 1.64, 2.33},
    xticklabels={1\%, 5\%, 10\%, 20\%, 50\%, 80\%, 90\%, 95\%, 99\%},
    ytick={-2.33, -1.64, -1.28, -0.84, 0, 0.84, 1.28, 1.64, 2.33},
    yticklabels={1\%, 5\%, 10\%, 20\%, 50\%, 80\%, 90\%, 95\%, 99\%},
    tick label style={font=\footnotesize},
]
\addlegendimage{draw, black}
\addlegendentry{BERT}
\addlegendimage{mark=triangle, only marks, black}
\addlegendentry{Haiku}
\addlegendimage{mark=triangle*, only marks, black}
\addlegendentry{Sonnet}
\addlegendimage{mark=square, only marks, black}
\addlegendentry{gpt-4o-mini}
\addlegendimage{mark=square*, only marks, black}
\addlegendentry{o4-mini}
\addlegendimage{dashed, gray}
\addlegendentry{Chance}

\addplot[care] table[x=fpr-probit-care, y=fnr-probit-care, col sep=comma] {data/classification_MFRC_DET.csv};
\addplot[mark=triangle, only marks, care, thick] coordinates {(-0.67, -0.39)}; 
\addplot[mark=triangle*, only marks, care] coordinates {(-0.55, -0.71)}; 
\addplot[mark=square, only marks, care, thick] coordinates {(-1.34, 0.00)}; 
\addplot[mark=square*, only marks, care] coordinates {(-0.67, -0.39)}; 

\addplot[loyalty] table[x=fpr-probit-loyalty, y=fnr-probit-loyalty, col sep=comma] {data/classification_MFRC_DET.csv};
\addplot[mark=triangle, only marks, loyalty, thick] coordinates {(-0.88, 0.27)}; 
\addplot[mark=triangle*, only marks, loyalty] coordinates {(-0.91, -0.14)}; 
\addplot[mark=square, only marks, loyalty, thick] coordinates {(-1.51, 0.80)}; 
\addplot[mark=square*, only marks, loyalty] coordinates {(-0.88, 0.27)}; 

\addplot[fairness] table[x=fpr-probit-fairness, y=fnr-probit-fairness, col sep=comma] {data/classification_MFRC_DET.csv};
\addplot[mark=triangle, only marks, fairness, thick] coordinates {(-0.15, -0.62)}; 
\addplot[mark=triangle*, only marks, fairness] coordinates {(-0.67, -0.18)}; 
\addplot[mark=square, only marks, fairness, thick] coordinates {(-1.00, 0.32)}; 
\addplot[mark=square*, only marks, fairness] coordinates {(-0.15, -0.62)}; 

\addplot[authority] table[x=fpr-probit-authority, y=fnr-probit-authority, col sep=comma] {data/classification_MFRC_DET.csv};
\addplot[mark=triangle, only marks, authority, thick] coordinates {(-0.81, 0.16)}; 
\addplot[mark=triangle*, only marks, authority] coordinates {(-0.76, -0.12)}; 
\addplot[mark=square, only marks, authority, thick] coordinates {(-1.29, 0.72)}; 
\addplot[mark=square*, only marks, authority] coordinates {(-0.81, 0.16)}; 

\addplot[sanctity, thick] table[x=fpr-probit-sanctity, y=fnr-probit-sanctity, col sep=comma] {data/classification_MFRC_DET.csv};
\addplot[mark=triangle, only marks, sanctity, thick] coordinates {(-1.51, 1.07)}; 
\addplot[mark=triangle*, only marks, sanctity] coordinates {(-1.20, 0.26)}; 
\addplot[mark=square, only marks, sanctity, thick] coordinates {(-1.80, 1.00)}; 
\addplot[mark=square*, only marks, sanctity] coordinates {(-1.51, 1.07)}; 

\addplot[gray, dashed] {-x};
\end{axis}
\end{tikzpicture}
\begin{center}
\small
\textcolor{care}{\rule{0.4cm}{2pt}} Care \quad  
\textcolor{fairness}{\rule{0.4cm}{2pt}} Fairness \quad
\textcolor{loyalty}{\rule{0.4cm}{2pt}} Loyalty \quad
\textcolor{authority}{\rule{0.4cm}{2pt}} Authority \quad
\textcolor{sanctity}{\rule{0.4cm}{2pt}} Sanctity
\end{center}
\caption{DET curves for MFRC dataset}
\label{fig:DET-MFRC-advanced}
\end{subfigure}\hfill
\begin{subfigure}{0.49\textwidth}
\centering
\begin{tikzpicture}
\begin{axis}[
   height=6cm,
   width=8cm,
    xlabel={False Positive Rate (Normal Deviate)},
    ylabel={False Negative Rate (Normal Deviate)},
    xmin=-2.33, xmax=1.64,
    ymin=-2.33, ymax=1.64,
    grid=major,
    legend pos=north east,
    legend cell align={left},
    legend style={font=\tiny, legend image post style={scale=0.8}, fill opacity=0.8},
    xtick={-2.33, -1.64, -1.28, -0.84, 0, 0.84, 1.28, 1.64, 2.33},
    xticklabels={1\%, 5\%, 10\%, 20\%, 50\%, 80\%, 90\%, 95\%, 99\%},
    ytick={-2.33, -1.64, -1.28, -0.84, 0, 0.84, 1.28, 1.64, 2.33},
    yticklabels={1\%, 5\%, 10\%, 20\%, 50\%, 80\%, 90\%, 95\%, 99\%},
    tick label style={font=\footnotesize},
]
\addlegendimage{draw, black}
\addlegendentry{BERT}
\addlegendimage{mark=triangle, only marks, black}
\addlegendentry{Haiku}
\addlegendimage{mark=triangle*, only marks, black}
\addlegendentry{Sonnet}
\addlegendimage{mark=square, only marks, black}
\addlegendentry{gpt-4o-mini}
\addlegendimage{mark=square*, only marks, black}
\addlegendentry{o4-mini}
\addlegendimage{dashed, gray}
\addlegendentry{Chance}

\addplot[care] table[x=fpr-probit-care, y=fnr-probit-care, col sep=comma] {data/classification_MFTC_DET.csv};
\addplot[mark=triangle, only marks, care, thick] coordinates {(-0.36, -0.45)}; 
\addplot[mark=triangle*, only marks, care] coordinates {(-0.31, -0.81)}; 
\addplot[mark=square, only marks, care, thick] coordinates {(-1.00, -0.08)}; 
\addplot[mark=square*, only marks, care] coordinates {(-0.55, -0.42)}; 

\addplot[loyalty] table[x=fpr-probit-loyalty, y=fnr-probit-loyalty, col sep=comma] {data/classification_MFTC_DET.csv};
\addplot[mark=triangle, only marks, loyalty, thick] coordinates {(-1.13, 0.53)}; 
\addplot[mark=triangle*, only marks, loyalty] coordinates {(-1.01, 0.28)}; 
\addplot[mark=square, only marks, loyalty, thick] coordinates {(-1.44, 0.63)}; 
\addplot[mark=square*, only marks, loyalty] coordinates {(-1.04, 0.29)}; 

\addplot[fairness] table[x=fpr-probit-fairness, y=fnr-probit-fairness, col sep=comma] {data/classification_MFTC_DET.csv};
\addplot[mark=triangle, only marks, fairness, thick] coordinates {(-0.55, -0.58)}; 
\addplot[mark=triangle*, only marks, fairness] coordinates {(-0.74, -0.52)}; 
\addplot[mark=square, only marks, fairness, thick] coordinates {(-1.08, 0.05)}; 
\addplot[mark=square*, only marks, fairness] coordinates {(-0.61, -0.55)}; 

\addplot[authority] table[x=fpr-probit-authority, y=fnr-probit-authority, col sep=comma] {data/classification_MFTC_DET.csv};
\addplot[mark=triangle, only marks, authority, thick] coordinates {(-1.08, 0.31)}; 
\addplot[mark=triangle*, only marks, authority] coordinates {(-0.96, 0.05)}; 
\addplot[mark=square, only marks, authority, thick] coordinates {(-1.60, 0.74)}; 
\addplot[mark=square*, only marks, authority] coordinates {(-0.97, 0.07)}; 

\addplot[sanctity, thick] table[x=fpr-probit-sanctity, y=fnr-probit-sanctity, col sep=comma] {data/classification_MFTC_DET.csv};
\addplot[mark=triangle, only marks, sanctity, thick] coordinates {(-1.59, 0.61)}; 
\addplot[mark=triangle*, only marks, sanctity] coordinates {(-0.76, -0.20)}; 
\addplot[mark=square, only marks, sanctity, thick] coordinates {(-1.88, 0.77)}; 
\addplot[mark=square*, only marks, sanctity] coordinates {(-1.50, 0.42)}; 

\addplot[gray, dashed] {-x};
\end{axis}
\end{tikzpicture}
\begin{center}
\small
\textcolor{care}{\rule{0.4cm}{2pt}} Care \quad  
\textcolor{fairness}{\rule{0.4cm}{2pt}} Fairness \quad
\textcolor{loyalty}{\rule{0.4cm}{2pt}} Loyalty \quad
\textcolor{authority}{\rule{0.4cm}{2pt}} Authority \quad
\textcolor{sanctity}{\rule{0.4cm}{2pt}} Sanctity
\end{center}
\caption{DET curves for MFTC dataset}
\label{fig:DET-MFTC-advanced}
\end{subfigure}
\label{fig:DET-advanced}
\end{figure*}

\subsubsection{Impact}

\begin{figure}
\centering
\pgfplotstableread{
Dataset Model GroundTruth Basic Engineered
MFRC Haiku-3.5 0.960313 0.983831 1.226604
MFRC Sonnet-3.5 0.960313 1.076923 2.040474
MFRC Sonnet-4 0.960313 1.041592 1.378354
MFRC gpt-4o-mini 0.960313 0.532053 0.670844
MFRC o4-mini 0.960313 0.469953 {}
MFRC {} {} {} {}
MFTC Haiku-3.5 1.627486 1.183485 1.321923
MFTC Sonnet-3.5 1.627486 1.355936 2.129829
MFTC Sonnet-4 1.627486 1.412168 1.739752
MFTC gpt-4o-mini 1.627486 0.735212 0.887246
MFTC o4-mini 1.627486 0.714083 1.355033
}\datatable
\resizebox{0.49\textwidth}{!}{%
\begin{tikzpicture}
\begin{axis}[
    width=12cm,
    height=8cm,
    xlabel={Model},
    ylabel={Label Cardinality},
    legend pos=outer north east,
    legend style={font=\small},
    legend entries={Basic, Engineered, Ground Truth},
    ybar=2pt,
    bar width=0.12cm,
    enlarge x limits=0.06,
    xtick=data,
    xticklabels from table={\datatable}{Model},
    xticklabel style={rotate=45, anchor=north east, font=\scriptsize},
    grid=major,
    grid style={dashed, gray!30},
    ymajorgrids=true,
    ymin=0,
    ymax=2.5,
    axis lines*=left,
    cycle list={
        {fill=gray!40, draw=gray!60},
        {fill=blue!60, draw=blue!80},
        {fill=black!80, draw=black!90}
    }
]
\addplot table[x expr=\coordindex, y=Basic] {\datatable};
\addplot table[x expr=\coordindex, y=Engineered] {\datatable};
\addplot table[x expr=\coordindex, y=GroundTruth] {\datatable};

\node[anchor=center, font=\large\bfseries] at (axis cs:2,2.35) {MFRC};
\node[anchor=center, font=\large\bfseries] at (axis cs:8.5,2.35) {MFTC};

\end{axis}
\end{tikzpicture}
}
\caption{Label cardinality by model and prompt type}
\label{fig:cardinality}
\end{figure}
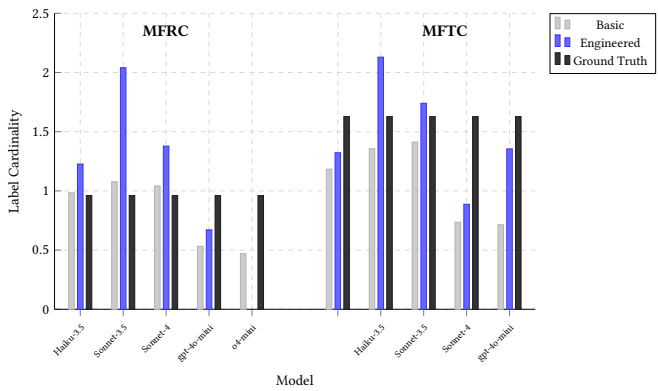

Performance curves under prompt engineering (Figures \subref{fig:ROC-MFRC-advanced}, \subref{fig:ROC-MFTC-advanced}, \subref{fig:PR-MFRC-advanced}, \subref{fig:PR-MFTC-advanced}, \subref{fig:DET-MFRC-advanced}, and \subref{fig:DET-MFTC-advanced}) confirm that enhanced prompting fails to close fundamental performance gaps. BERT maintains superior AUC and dominates the precision-recall space, while LLM curves remain largely within the transformer envelope with inconsistent gains and unchanged high false negative rate patterns.

\subsection{Under-Detection of Moral Content}

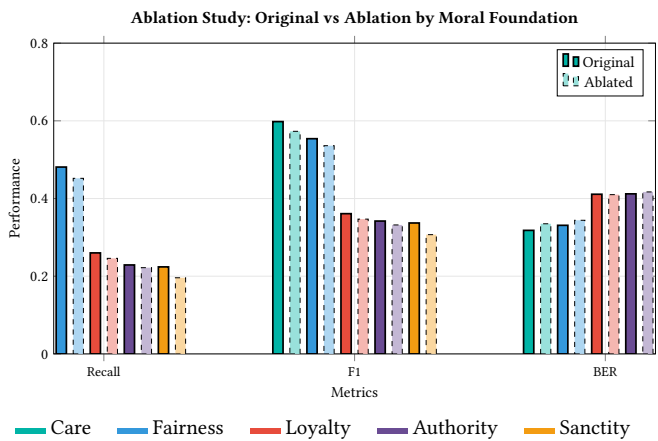
\begin{figure}
\resizebox{0.49\textwidth}{!}{%
\begin{tikzpicture}
\begin{axis}[
    width=14cm,
    height=8cm,
    xlabel={Metrics},
    ylabel={Performance},
    symbolic x coords={Recall, F1, BER},
    xtick=data,
    x tick label style={font=\small},
    ybar=4pt,
    bar width=6pt,
    grid=major,
    grid style={gray!20},
    ymin=0,
    ymax=0.8,
    title={Ablation Study: Original vs Ablation by Moral Foundation},
    title style={font=\large\bfseries},
]
\addplot[fill=care, draw=black, line width=1pt] coordinates {
    (Recall, 0.527)
    (F1, 0.598)
    (BER, 0.318)
};
\addplot[fill=care!40, draw=black, line width=0.5pt, dashed] coordinates {
    (Recall, 0.489)
    (F1, 0.573)
    (BER, 0.335)
};
\addplot[fill=fairness, draw=black, line width=1pt] coordinates {
    (Recall, 0.481)
    (F1, 0.554)
    (BER, 0.331)
};
\addplot[fill=fairness!40, draw=black, line width=0.5pt, dashed] coordinates {
    (Recall, 0.452)
    (F1, 0.536)
    (BER, 0.344)
};
\addplot[fill=loyalty, draw=black, line width=1pt] coordinates {
    (Recall, 0.260)
    (F1, 0.361)
    (BER, 0.411)
};
\addplot[fill=loyalty!40, draw=black, line width=0.5pt, dashed] coordinates {
    (Recall, 0.246)
    (F1, 0.347)
    (BER, 0.410)
};
\addplot[fill=authority, draw=black, line width=1pt] coordinates {
    (Recall, 0.229)
    (F1, 0.342)
    (BER, 0.412)
};
\addplot[fill=authority!40, draw=black, line width=0.5pt, dashed] coordinates {
    (Recall, 0.222)
    (F1, 0.332)
    (BER, 0.417)
};
\addplot[fill=sanctity, draw=black, line width=1pt] coordinates {
    (Recall, 0.224)
    (F1, 0.337)
    (BER, 0.404)
};
\addplot[fill=sanctity!40, draw=black, line width=0.5pt, dashed] coordinates {
    (Recall, 0.196)
    (F1, 0.307)
    (BER, 0.411)
};
\legend{
    Original,
    Ablated
}
\end{axis}
\end{tikzpicture}
}
\begin{center}
\small
\textcolor{care}{\rule{0.4cm}{2pt}} Care \quad  
\textcolor{fairness}{\rule{0.4cm}{2pt}} Fairness \quad
\textcolor{loyalty}{\rule{0.4cm}{2pt}} Loyalty \quad
\textcolor{authority}{\rule{0.4cm}{2pt}} Authority \quad
\textcolor{sanctity}{\rule{0.4cm}{2pt}} Sanctity
\end{center}
\caption{Impact of Removed Recall-Improving Instructions}
\label{fig:ablation}
\end{figure}

Our ablation study reveals fundamental limitations in LLMs' moral reasoning approach. \Cref{fig:ablation} demonstrates that LLMs require explicit recall-increasing instructions to identify moral foundations, suggesting they inherently under-detect moral content. Even with enhanced prompting that reinforces multi-label classification requirements, \Cref{fig:cardinality} shows LLMs consistently predict lower label cardinality than ground truth, indicating systematic under-prediction of moral dimensions. This conservative bias explains their high false negative rates and suggests that LLMs lack the nuanced understanding necessary to recognize the complex, overlapping nature of moral foundations in text.

\section{CONCLUSION}

This study presents the first comprehensive comparison between fine-tuned transformer models and state-of-the-art LLMs for moral foundation classification. Despite their flexibility and accessibility, LLMs significantly underperform across multiple evaluation metrics, exhibiting excessive false negatives on complex moral dimensions such as loyalty and sanctity. Fine-tuned transformers such as DeBERTa and RoBERTa outperform LLMs by substantial margins, particularly on nuanced foundation-level classification. Critically, prompt engineering yielded inconsistent and minimal improvements, with some models actually degrading, demonstrating that prompting general-purpose LLMs cannot reliably substitute for task-specific fine-tuning. These findings reveal fundamental limitations in LLMs' moral reasoning capabilities and systematic under-detection of moral content.

Given LLMs' increasing integration into ethically sensitive applications—chatbots, content moderation, decision aids—these accuracy limitations raise serious deployment concerns. The AI community must prioritize specialized models for moral analysis, implement mandatory human oversight for LLM moral applications, and develop hybrid approaches combining LLM accessibility with transformer precision. Future research should focus on culturally diverse moral datasets, systematic error analysis, and embedding explicit moral reasoning into training protocols rather than relying on post-hoc prompting strategies.

Bridging the gap between technical performance and moral responsibility requires acknowledging current limitations and investing in purpose-built solutions—ensuring language technologies serve society both intelligently and ethically.

\bibliographystyle{apalike}
\bibliography{citations}

\end{document}

%% file: classification.tex

The results of our evaluation, presented in \Cref{tab:all_results_MFRC} and \Cref{tab:all_results_MFTC}, reveal substantial performance gaps between fine-tuned transformers and LLMs. BERT consistently outperforms all LLMs with ROC-AUC scores of 0.83-0.90 and F1 scores of 0.38-0.80, while Claude-3.5-Sonnet achieves only 0.21-0.78 F1. LLMs struggle most with loyalty and sanctity, exhibiting high false negative rates (0.58-0.90) that miss most positive cases. While LLMs detect general moral content reasonably well, their inability to identify specific foundations demonstrates that task-specific fine-tuning remains superior to prompting for moral reasoning tasks.


\begin{table}[h]
\centering
\caption{Performance metrics for MFRC under simple prompt}
\label{tab:all_results_MFRC}
\resizebox{0.8\columnwidth}{!}{
\begin{tabular}{lllrrrrrrr}
\toprule
 &  &  & FPR & FNR & precision & recall & f1 & ROC-AUC & AP \\
dataset & moral dimension & model &  &  &  &  &  &  &  \\
\midrule
\multirow[t]{42}{*}{MFRC} & \multirow[t]{7}{*}{any} & BERT & 0.225 & 0.221 & 0.761 & 0.862 & 0.808 & 0.862 & 0.893 \\
 &  & BERT-OOD & 0.247 & 0.306 & 0.691 & 0.892 & 0.779 & 0.800 & 0.829 \\
 &  & Haiku 3.5 & 0.469 & 0.180 & 0.696 & 0.820 & 0.753 &  &  \\
 &  & Sonnet 3.7 & 0.374 & 0.184 & 0.741 & 0.816 & 0.777 &  &  \\
 &  & Sonnet 4 & 0.369 & 0.178 & 0.745 & 0.822 & 0.782 &  &  \\
 &  & gpt-4o-mini & 0.243 & 0.352 & 0.778 & 0.648 & 0.707 &  &  \\
 &  & o4-mini & 0.195 & 0.412 & 0.798 & 0.588 & 0.677 &  &  \\
\cline{2-10}
 & \multirow[t]{7}{*}{authority} & BERT & 0.276 & 0.137 & 0.549 & 0.638 & 0.590 & 0.869 & 0.605 \\
 &  & BERT-OOD & 0.280 & 0.297 & 0.422 & 0.600 & 0.495 & 0.775 & 0.453 \\
 &  & Haiku 3.5 & 0.229 & 0.529 & 0.330 & 0.471 & 0.388 &  &  \\
 &  & Sonnet 3.7 & 0.143 & 0.523 & 0.444 & 0.477 & 0.460 &  &  \\
 &  & Sonnet 4 & 0.144 & 0.573 & 0.416 & 0.427 & 0.421 &  &  \\
 &  & gpt-4o-mini & 0.146 & 0.623 & 0.383 & 0.377 & 0.380 &  &  \\
 &  & o4-mini & 0.091 & 0.769 & 0.379 & 0.231 & 0.287 &  &  \\
\cline{2-10}
 & \multirow[t]{7}{*}{care} & BERT & 0.217 & 0.156 & 0.691 & 0.720 & 0.706 & 0.896 & 0.775 \\
 &  & BERT-OOD & 0.265 & 0.234 & 0.533 & 0.727 & 0.615 & 0.823 & 0.626 \\
 &  & Haiku 3.5 & 0.255 & 0.405 & 0.451 & 0.595 & 0.513 &  &  \\
 &  & Sonnet 3.7 & 0.155 & 0.359 & 0.593 & 0.641 & 0.616 &  &  \\
 &  & Sonnet 4 & 0.192 & 0.316 & 0.556 & 0.684 & 0.613 &  &  \\
 &  & gpt-4o-mini & 0.137 & 0.477 & 0.573 & 0.523 & 0.547 &  &  \\
 &  & o4-mini & 0.155 & 0.471 & 0.546 & 0.529 & 0.537 &  &  \\
\cline{2-10}
 & \multirow[t]{7}{*}{fairness} & BERT & 0.275 & 0.215 & 0.544 & 0.785 & 0.643 & 0.830 & 0.711 \\
 &  & BERT-OOD & 0.368 & 0.266 & 0.455 & 0.734 & 0.562 & 0.745 & 0.533 \\
 &  & Haiku 3.5 & 0.362 & 0.416 & 0.419 & 0.584 & 0.488 &  &  \\
 &  & Sonnet 3.7 & 0.264 & 0.387 & 0.509 & 0.613 & 0.556 &  &  \\
 &  & Sonnet 4 & 0.201 & 0.493 & 0.530 & 0.507 & 0.518 &  &  \\
 &  & gpt-4o-mini & 0.125 & 0.765 & 0.456 & 0.235 & 0.310 &  &  \\
 &  & o4-mini & 0.150 & 0.618 & 0.532 & 0.382 & 0.445 &  &  \\
\cline{2-10}
 & \multirow[t]{7}{*}{loyalty} & BERT & 0.194 & 0.202 & 0.471 & 0.639 & 0.542 & 0.878 & 0.563 \\
 &  & BERT-OOD & 0.243 & 0.357 & 0.372 & 0.415 & 0.392 & 0.777 & 0.395 \\
 &  & Haiku 3.5 & 0.150 & 0.685 & 0.199 & 0.315 & 0.244 &  &  \\
 &  & Sonnet 3.7 & 0.091 & 0.584 & 0.351 & 0.416 & 0.381 &  &  \\
 &  & Sonnet 4 & 0.100 & 0.535 & 0.354 & 0.465 & 0.402 &  &  \\
 &  & gpt-4o-mini & 0.052 & 0.892 & 0.198 & 0.108 & 0.140 &  &  \\
 &  & o4-mini & 0.072 & 0.801 & 0.245 & 0.199 & 0.220 &  &  \\
\cline{2-10}
 & \multirow[t]{7}{*}{sanctity} & BERT & 0.293 & 0.131 & 0.358 & 0.566 & 0.439 & 0.859 & 0.444 \\
 &  & BERT-OOD & 0.283 & 0.258 & 0.305 & 0.531 & 0.388 & 0.804 & 0.376 \\
 &  & Haiku 3.5 & 0.160 & 0.715 & 0.171 & 0.285 & 0.213 &  &  \\
 &  & Sonnet 3.7 & 0.067 & 0.695 & 0.345 & 0.305 & 0.323 &  &  \\
 &  & Sonnet 4 & 0.064 & 0.695 & 0.353 & 0.305 & 0.327 &  &  \\
 &  & gpt-4o-mini & 0.060 & 0.903 & 0.157 & 0.097 & 0.120 &  &  \\
 &  & o4-mini & 0.076 & 0.857 & 0.178 & 0.143 & 0.158 &  &  \\
\cline{1-10} \cline{2-10}
\bottomrule
\end{tabular}

}
\end{table}

\begin{table}
    \caption{Performance metrics on MFTC under simple prompt}
    \label{tab:all_results_MFTC}
    \resizebox{0.8\columnwidth}{!}{
\begin{tabular}{lllrrrrrrr}
\toprule
 &  &  & FPR & FNR & precision & recall & f1 & ROC-AUC & AP \\
dataset & moral dimension & model &  &  &  &  &  &  &  \\
\midrule
\multirow[t]{42}{*}{MFTC} & \multirow[t]{7}{*}{any} & BERT & 0.075 & 0.176 & 0.902 & 0.959 & 0.930 & 0.940 & 0.983 \\
 &  & BERT-OOD & 0.189 & 0.238 & 0.867 & 0.934 & 0.899 & 0.860 & 0.957 \\
 &  & Haiku 3.5 & 0.372 & 0.158 & 0.896 & 0.842 & 0.868 &  &  \\
 &  & Sonnet 3.7 & 0.321 & 0.155 & 0.909 & 0.845 & 0.876 &  &  \\
 &  & Sonnet 4 & 0.328 & 0.152 & 0.908 & 0.848 & 0.877 &  &  \\
 &  & gpt-4o-mini & 0.291 & 0.287 & 0.903 & 0.713 & 0.797 &  &  \\
 &  & o4-mini & 0.204 & 0.305 & 0.929 & 0.695 & 0.795 &  &  \\
\cline{2-10}
 & \multirow[t]{7}{*}{authority} & BERT & 0.204 & 0.173 & 0.743 & 0.742 & 0.742 & 0.899 & 0.830 \\
 &  & BERT-OOD & 0.205 & 0.395 & 0.567 & 0.645 & 0.603 & 0.749 & 0.645 \\
 &  & Haiku 3.5 & 0.151 & 0.611 & 0.551 & 0.389 & 0.456 &  &  \\
 &  & Sonnet 3.7 & 0.136 & 0.553 & 0.610 & 0.447 & 0.516 &  &  \\
 &  & Sonnet 4 & 0.139 & 0.583 & 0.589 & 0.417 & 0.489 &  &  \\
 &  & gpt-4o-mini & 0.201 & 0.603 & 0.485 & 0.397 & 0.436 &  &  \\
 &  & o4-mini & 0.146 & 0.698 & 0.498 & 0.302 & 0.376 &  &  \\
\cline{2-10}
 & \multirow[t]{7}{*}{care} & BERT & 0.115 & 0.244 & 0.805 & 0.769 & 0.787 & 0.897 & 0.881 \\
 &  & BERT-OOD & 0.252 & 0.261 & 0.663 & 0.748 & 0.703 & 0.811 & 0.765 \\
 &  & Haiku 3.5 & 0.314 & 0.343 & 0.594 & 0.657 & 0.624 &  &  \\
 &  & Sonnet 3.7 & 0.295 & 0.298 & 0.624 & 0.702 & 0.661 &  &  \\
 &  & Sonnet 4 & 0.339 & 0.254 & 0.606 & 0.746 & 0.668 &  &  \\
 &  & gpt-4o-mini & 0.251 & 0.447 & 0.606 & 0.553 & 0.579 &  &  \\
 &  & o4-mini & 0.224 & 0.400 & 0.651 & 0.600 & 0.624 &  &  \\
\cline{2-10}
 & \multirow[t]{7}{*}{fairness} & BERT & 0.111 & 0.232 & 0.792 & 0.768 & 0.780 & 0.906 & 0.874 \\
 &  & BERT-OOD & 0.317 & 0.286 & 0.512 & 0.819 & 0.630 & 0.770 & 0.659 \\
 &  & Haiku 3.5 & 0.243 & 0.338 & 0.601 & 0.662 & 0.630 &  &  \\
 &  & Sonnet 3.7 & 0.204 & 0.287 & 0.659 & 0.713 & 0.685 &  &  \\
 &  & Sonnet 4 & 0.197 & 0.322 & 0.655 & 0.678 & 0.666 &  &  \\
 &  & gpt-4o-mini & 0.128 & 0.734 & 0.534 & 0.266 & 0.355 &  &  \\
 &  & o4-mini & 0.190 & 0.437 & 0.620 & 0.563 & 0.590 &  &  \\
\cline{2-10}
 & \multirow[t]{7}{*}{loyalty} & BERT & 0.159 & 0.267 & 0.670 & 0.733 & 0.700 & 0.866 & 0.782 \\
 &  & BERT-OOD & 0.228 & 0.460 & 0.442 & 0.680 & 0.536 & 0.713 & 0.551 \\
 &  & Haiku 3.5 & 0.083 & 0.818 & 0.485 & 0.182 & 0.265 &  &  \\
 &  & Sonnet 3.7 & 0.071 & 0.711 & 0.636 & 0.289 & 0.398 &  &  \\
 &  & Sonnet 4 & 0.083 & 0.676 & 0.627 & 0.324 & 0.427 &  &  \\
 &  & gpt-4o-mini & 0.094 & 0.880 & 0.355 & 0.120 & 0.179 &  &  \\
 &  & o4-mini & 0.124 & 0.786 & 0.425 & 0.214 & 0.284 &  &  \\
\cline{2-10}
 & \multirow[t]{7}{*}{sanctity} & BERT & 0.138 & 0.262 & 0.688 & 0.654 & 0.671 & 0.875 & 0.740 \\
 &  & BERT-OOD & 0.340 & 0.271 & 0.433 & 0.616 & 0.509 & 0.761 & 0.523 \\
 &  & Haiku 3.5 & 0.122 & 0.762 & 0.360 & 0.238 & 0.286 &  &  \\
 &  & Sonnet 3.7 & 0.080 & 0.674 & 0.541 & 0.326 & 0.407 &  &  \\
 &  & Sonnet 4 & 0.095 & 0.646 & 0.517 & 0.354 & 0.420 &  &  \\
 &  & gpt-4o-mini & 0.097 & 0.839 & 0.325 & 0.161 & 0.216 &  &  \\
 &  & o4-mini & 0.115 & 0.782 & 0.354 & 0.218 & 0.270 &  &  \\
\cline{1-10} \cline{2-10}
\bottomrule
\end{tabular}

}
\end{table}

\subsubsection{ROC Analysis}

Figures \subref{fig:ROC-MFRC-simple} and \subref{fig:ROC-MFTC-simple} show ROC curves plotting true positive rate against false positive rate across all classification thresholds. The curves visually confirm the substantial performance gaps reported in the tables, with BERT consistently achieving higher AUC values while LLM curves lie systematically inside the transformer performance envelope.

\begin{figure*}[h!]
\centering
\begin{subfigure}{0.49\textwidth}
\centering
\begin{tikzpicture}
\begin{axis}[
   height=6cm,
   width=8cm,
   xlabel={False Positive Rate},
   ylabel={True Positive Rate},
   xmin=0, xmax=1,
   ymin=0, ymax=1,
   grid=major,
   legend pos=south east,
   legend cell align={left},
   legend style={font=\tiny, legend image post style={scale=0.8}, fill opacity=0.8,}
]

\addlegendimage{draw, black}
\addlegendentry{BERT}
\addlegendimage{mark=triangle, only marks, black}
\addlegendentry{Haiku}
\addlegendimage{mark=triangle*, only marks, black}
\addlegendentry{Sonnet}
\addlegendimage{mark=square, only marks, black}
\addlegendentry{gpt-4o-mini}
\addlegendimage{mark=square*, only marks, black}
\addlegendentry{o4-mini}
\addlegendimage{dashed, gray}
\addlegendentry{Chance}

\addplot[care] table[x=fpr-care, y=tpr-care, col sep=comma] {data/classification_MFRC_ROC.csv};
\addplot[mark=triangle, only marks, care, thick] coordinates {(0.26,1-0.41)}; 
\addplot[mark=triangle*, only marks, care] coordinates {(0.16,1-0.36)}; 
\addplot[mark=square, only marks, care, thick] coordinates {(0.14,1-0.48)}; 
\addplot[mark=square*, only marks, care] coordinates {(0.16,1-0.47)}; 

\addplot[loyalty] table[x=fpr-loyalty, y=tpr-loyalty, col sep=comma] {data/classification_MFRC_ROC.csv};
\addplot[mark=triangle, only marks, loyalty, thick] coordinates {(0.15,1-0.68)}; 
\addplot[mark=triangle*, only marks, loyalty] coordinates {(0.09,1-0.58)}; 
\addplot[mark=square, only marks, loyalty, thick] coordinates {(0.05,1-0.89)}; 
\addplot[mark=square*, only marks, loyalty] coordinates {(0.07,1-0.80)}; 

\addplot[fairness] table[x=fpr-fairness, y=tpr-fairness, col sep=comma] {data/classification_MFRC_ROC.csv};
\addplot[mark=triangle, only marks, fairness, thick] coordinates {(0.36,1-0.42)}; 
\addplot[mark=triangle*, only marks, fairness] coordinates {(0.26,1-0.39)}; 
\addplot[mark=square, only marks, fairness, thick] coordinates {(0.13,1-0.77)}; 
\addplot[mark=square*, only marks, fairness] coordinates {(0.15,1-0.62)}; 

\addplot[authority] table[x=fpr-authority, y=tpr-authority, col sep=comma] {data/classification_MFRC_ROC.csv};
\addplot[mark=triangle, only marks, authority, thick] coordinates {(0.23,1-0.53)}; 
\addplot[mark=triangle*, only marks, authority] coordinates {(0.14,1-0.52)}; 
\addplot[mark=square, only marks, authority, thick] coordinates {(0.15,1-0.62)}; 
\addplot[mark=square*, only marks, authority] coordinates {(0.09,1-0.77)}; 

\addplot[sanctity, thick] table[x=fpr-sanctity, y=tpr-sanctity, col sep=comma] {data/classification_MFRC_ROC.csv};
\addplot[mark=triangle, only marks, sanctity, thick] coordinates {(0.16,1-0.72)}; 
\addplot[mark=triangle*, only marks, sanctity] coordinates {(0.07,1-0.70)}; 
\addplot[mark=square, only marks, sanctity, thick] coordinates {(0.06,1-0.90)}; 
\addplot[mark=square*, only marks, sanctity] coordinates {(0.08,1-0.86)}; 

\addplot[dashed, gray] coordinates {(0,0) (1,1)};

\end{axis}
\end{tikzpicture}
\begin{center}
\small
\textcolor{care}{\rule{0.4cm}{2pt}} Care \quad  
\textcolor{fairness}{\rule{0.4cm}{2pt}} Fairness \quad
\textcolor{loyalty}{\rule{0.4cm}{2pt}} Loyalty \quad
\textcolor{authority}{\rule{0.4cm}{2pt}} Authority \quad
\textcolor{sanctity}{\rule{0.4cm}{2pt}} Sanctity
\end{center}
\caption{ROC curves for MFRC dataset}
\label{fig:ROC-MFRC-simple}
\end{subfigure}\hfill
\begin{subfigure}{0.49\textwidth}
\centering
\begin{tikzpicture}
\begin{axis}[
   height=6cm,
   width=8cm,
   xlabel={False Positive Rate},
   ylabel={True Positive Rate},
   xmin=0, xmax=1,
   ymin=0, ymax=1,
   grid=major,
   legend pos=south east,
   legend cell align={left},
   legend style={font=\tiny, legend image post style={scale=0.8}}
]

\addlegendimage{draw, black}
\addlegendentry{BERT}
\addlegendimage{mark=triangle, only marks, black}
\addlegendentry{Haiku}
\addlegendimage{mark=triangle*, only marks, black}
\addlegendentry{Sonnet}
\addlegendimage{mark=square, only marks, black}
\addlegendentry{gpt-4o-mini}
\addlegendimage{mark=square*, only marks, black}
\addlegendentry{o4-mini}
\addlegendimage{dashed, gray}
\addlegendentry{Chance}

\addplot[care] table[x=fpr-care, y=tpr-care, col sep=comma] {data/classification_MFTC_ROC.csv};
\addplot[mark=triangle, only marks, care, thick] coordinates {(0.31,1-0.34)}; 
\addplot[mark=triangle*, only marks, care] coordinates {(0.30,1-0.30)}; 
\addplot[mark=square, only marks, care, thick] coordinates {(0.25,1-0.45)}; 
\addplot[mark=square*, only marks, care] coordinates {(0.22,1-0.40)}; 

\addplot[loyalty] table[x=fpr-loyalty, y=tpr-loyalty, col sep=comma] {data/classification_MFTC_ROC.csv};
\addplot[mark=triangle, only marks, loyalty, thick] coordinates {(0.08,1-0.82)}; 
\addplot[mark=triangle*, only marks, loyalty] coordinates {(0.07,1-0.71)}; 
\addplot[mark=square, only marks, loyalty, thick] coordinates {(0.09,1-0.88)}; 
\addplot[mark=square*, only marks, loyalty] coordinates {(0.12,1-0.79)}; 

\addplot[fairness] table[x=fpr-fairness, y=tpr-fairness, col sep=comma] {data/classification_MFTC_ROC.csv};
\addplot[mark=triangle, only marks, fairness, thick] coordinates {(0.24,1-0.34)}; 
\addplot[mark=triangle*, only marks, fairness] coordinates {(0.20,1-0.29)}; 
\addplot[mark=square, only marks, fairness, thick] coordinates {(0.13,1-0.73)}; 
\addplot[mark=square*, only marks, fairness] coordinates {(0.19,1-0.44)}; 

\addplot[authority] table[x=fpr-authority, y=tpr-authority, col sep=comma] {data/classification_MFTC_ROC.csv};
\addplot[mark=triangle, only marks, authority, thick] coordinates {(0.15,1-0.61)}; 
\addplot[mark=triangle*, only marks, authority] coordinates {(0.14,1-0.55)}; 
\addplot[mark=square, only marks, authority, thick] coordinates {(0.20,1-0.60)}; 
\addplot[mark=square*, only marks, authority] coordinates {(0.15,1-0.70)}; 

\addplot[sanctity, thick] table[x=fpr-sanctity, y=tpr-sanctity, col sep=comma] {data/classification_MFTC_ROC.csv};
\addplot[mark=triangle, only marks, sanctity, thick] coordinates {(0.12,1-0.76)}; 
\addplot[mark=triangle*, only marks, sanctity] coordinates {(0.08,1-0.67)}; 
\addplot[mark=square, only marks, sanctity, thick] coordinates {(0.10,1-0.84)}; 
\addplot[mark=square*, only marks, sanctity] coordinates {(0.11,1-0.78)}; 

\addplot[dashed, gray] coordinates {(0,0) (1,1)};

\end{axis}
\end{tikzpicture}
\begin{center}
\small
\textcolor{care}{\rule{0.4cm}{2pt}} Care \quad  
\textcolor{fairness}{\rule{0.4cm}{2pt}} Fairness \quad
\textcolor{loyalty}{\rule{0.4cm}{2pt}} Loyalty \quad
\textcolor{authority}{\rule{0.4cm}{2pt}} Authority \quad
\textcolor{sanctity}{\rule{0.4cm}{2pt}} Sanctity
\end{center}
\caption{ROC curves for MFTC dataset}
\label{fig:ROC-MFTC-simple}
\end{subfigure}
\label{fig:ROC-simple}
\end{figure*}

\subsubsection{PR Analysis}

Figures \subref{fig:PR-MFRC-simple} and \subref{fig:PR-MFTC-simple} present precision-recall curves, which are particularly informative for imbalanced datasets like moral foundation classification. These plots corroborate the tabular results, demonstrating that fine-tuned transformers maintain superior precision-recall trade-offs compared to LLMs across all moral foundations.

\begin{figure*}[h!]
\centering
\begin{subfigure}{0.49\textwidth}
\centering
\begin{tikzpicture}
\begin{axis}[
   height=6cm,
   width=8cm,
   xlabel={Recall},
   ylabel={Precision},
   xmin=0, xmax=1,
   ymin=0, ymax=1,
   grid=major,
   legend pos=north east,
   legend cell align={left},
   legend style={font=\tiny, legend image post style={scale=0.8}, fill opacity=0.6,}
]
\addlegendimage{draw, black}
\addlegendentry{BERT}
\addlegendimage{mark=triangle, only marks, black}
\addlegendentry{Haiku}
\addlegendimage{mark=triangle*, only marks, black}
\addlegendentry{Sonnet}
\addlegendimage{mark=square, only marks, black}
\addlegendentry{gpt-4o-mini}
\addlegendimage{mark=square*, only marks, black}
\addlegendentry{o4-mini}

\addplot[care] table[x=recall-care, y=precision-care, col sep=comma] {data/classification_MFRC_PR.csv};
\addplot[mark=triangle, only marks, care, thick] coordinates {(0.60,0.45)}; 
\addplot[mark=triangle*, only marks, care] coordinates {(0.64,0.59)}; 
\addplot[mark=square, only marks, care, thick] coordinates {(0.52,0.57)}; 
\addplot[mark=square*, only marks, care] coordinates {(0.53,0.55)}; 

\addplot[loyalty] table[x=recall-loyalty, y=precision-loyalty, col sep=comma] {data/classification_MFRC_PR.csv};
\addplot[mark=triangle, only marks, loyalty, thick] coordinates {(0.32,0.20)}; 
\addplot[mark=triangle*, only marks, loyalty] coordinates {(0.42,0.35)}; 
\addplot[mark=square, only marks, loyalty, thick] coordinates {(0.11,0.20)}; 
\addplot[mark=square*, only marks, loyalty] coordinates {(0.20,0.25)}; 

\addplot[fairness] table[x=recall-fairness, y=precision-fairness, col sep=comma] {data/classification_MFRC_PR.csv};
\addplot[mark=triangle, only marks, fairness, thick] coordinates {(0.58,0.42)}; 
\addplot[mark=triangle*, only marks, fairness] coordinates {(0.61,0.51)}; 
\addplot[mark=square, only marks, fairness, thick] coordinates {(0.23,0.46)}; 
\addplot[mark=square*, only marks, fairness] coordinates {(0.38,0.53)}; 

\addplot[authority] table[x=recall-authority, y=precision-authority, col sep=comma] {data/classification_MFRC_PR.csv};
\addplot[mark=triangle, only marks, authority, thick] coordinates {(0.47,0.33)}; 
\addplot[mark=triangle*, only marks, authority] coordinates {(0.48,0.44)}; 
\addplot[mark=square, only marks, authority, thick] coordinates {(0.38,0.38)}; 
\addplot[mark=square*, only marks, authority] coordinates {(0.23,0.38)}; 

\addplot[sanctity, thick] table[x=recall-sanctity, y=precision-sanctity, col sep=comma] {data/classification_MFRC_PR.csv};
\addplot[mark=triangle, only marks, sanctity, thick] coordinates {(0.29,0.17)}; 
\addplot[mark=triangle*, only marks, sanctity] coordinates {(0.30,0.34)}; 
\addplot[mark=square, only marks, sanctity, thick] coordinates {(0.10,0.16)}; 
\addplot[mark=square*, only marks, sanctity] coordinates {(0.14,0.18)}; 

\end{axis}
\end{tikzpicture}
\begin{center}
\small
\textcolor{care}{\rule{0.4cm}{2pt}} Care \quad  
\textcolor{fairness}{\rule{0.4cm}{2pt}} Fairness \quad
\textcolor{loyalty}{\rule{0.4cm}{2pt}} Loyalty \quad
\textcolor{authority}{\rule{0.4cm}{2pt}} Authority \quad
\textcolor{sanctity}{\rule{0.4cm}{2pt}} Sanctity
\end{center}
\caption{Precision-recall curves for MFRC dataset}
\label{fig:PR-MFRC-simple}
\end{subfigure}\hfill
\begin{subfigure}{0.49\textwidth}
\centering
\begin{tikzpicture}
\begin{axis}[
   height=6cm,
   width=8cm,
   xlabel={Recall},
   ylabel={Precision},
   xmin=0, xmax=1,
   ymin=0, ymax=1,
   grid=major,
   legend pos=north east,
   legend cell align={left},
   legend style={font=\tiny, legend image post style={scale=0.8}, fill opacity=0.8,}
]
\addlegendimage{draw, black}
\addlegendentry{BERT}
\addlegendimage{mark=triangle, only marks, black}
\addlegendentry{Haiku}
\addlegendimage{mark=triangle*, only marks, black}
\addlegendentry{Sonnet}
\addlegendimage{mark=square, only marks, black}
\addlegendentry{gpt-4o-mini}
\addlegendimage{mark=square*, only marks, black}
\addlegendentry{o4-mini}

\addplot[care] table[x=recall-care, y=precision-care, col sep=comma] {data/classification_MFTC_PR.csv};
\addplot[mark=triangle, only marks, care, thick] coordinates {(0.66,0.59)}; 
\addplot[mark=triangle*, only marks, care] coordinates {(0.70,0.62)}; 
\addplot[mark=square, only marks, care, thick] coordinates {(0.55,0.61)}; 
\addplot[mark=square*, only marks, care] coordinates {(0.60,0.65)}; 

\addplot[loyalty] table[x=recall-loyalty, y=precision-loyalty, col sep=comma] {data/classification_MFTC_PR.csv};
\addplot[mark=triangle, only marks, loyalty, thick] coordinates {(0.18,0.48)}; 
\addplot[mark=triangle*, only marks, loyalty] coordinates {(0.29,0.64)}; 
\addplot[mark=square, only marks, loyalty, thick] coordinates {(0.12,0.35)}; 
\addplot[mark=square*, only marks, loyalty] coordinates {(0.21,0.43)}; 

\addplot[fairness] table[x=recall-fairness, y=precision-fairness, col sep=comma] {data/classification_MFTC_PR.csv};
\addplot[mark=triangle, only marks, fairness, thick] coordinates {(0.66,0.60)}; 
\addplot[mark=triangle*, only marks, fairness] coordinates {(0.71,0.66)}; 
\addplot[mark=square, only marks, fairness, thick] coordinates {(0.27,0.53)}; 
\addplot[mark=square*, only marks, fairness] coordinates {(0.56,0.62)}; 

\addplot[authority] table[x=recall-authority, y=precision-authority, col sep=comma] {data/classification_MFTC_PR.csv};
\addplot[mark=triangle, only marks, authority, thick] coordinates {(0.39,0.55)}; 
\addplot[mark=triangle*, only marks, authority] coordinates {(0.45,0.61)}; 
\addplot[mark=square, only marks, authority, thick] coordinates {(0.40,0.48)}; 
\addplot[mark=square*, only marks, authority] coordinates {(0.30,0.50)}; 

\addplot[sanctity, thick] table[x=recall-sanctity, y=precision-sanctity, col sep=comma] {data/classification_MFTC_PR.csv};
\addplot[mark=triangle, only marks, sanctity, thick] coordinates {(0.24,0.36)}; 
\addplot[mark=triangle*, only marks, sanctity] coordinates {(0.33,0.54)}; 
\addplot[mark=square, only marks, sanctity, thick] coordinates {(0.16,0.33)}; 
\addplot[mark=square*, only marks, sanctity] coordinates {(0.22,0.35)}; 

\end{axis}
\end{tikzpicture}
\begin{center}
\begin{center}
\small
\textcolor{care}{\rule{0.4cm}{2pt}} Care \quad  
\textcolor{fairness}{\rule{0.4cm}{2pt}} Fairness \quad
\textcolor{loyalty}{\rule{0.4cm}{2pt}} Loyalty \quad
\textcolor{authority}{\rule{0.4cm}{2pt}} Authority \quad
\textcolor{sanctity}{\rule{0.4cm}{2pt}} Sanctity
\end{center}
\end{center}
\caption{Precision-recall curves for MFTC dataset}
\label{fig:PR-MFTC-simple}
\end{subfigure}
\end{figure*}

\subsubsection{DET Curves}

Results in \subref{fig:DET-MFRC-simple} and \subref{fig:DET-MFTC-simple} show Detection Error Tradeoff curves plotting false negative against false positive rates on normal deviate scales. These curves emphasize the error rate perspective and confirm the high false negative rates of LLMs documented in the tables, particularly for loyalty and sanctity foundations.

\begin{figure*}
\centering
\begin{subfigure}{0.49\textwidth}
\centering
\begin{tikzpicture}
\begin{axis}[
   height=6cm,
   width=8cm,
    xlabel={False Positive Rate (Normal Deviate)},
    ylabel={False Negative Rate (Normal Deviate)},
    xmin=-2.33, xmax=1.64, 
    ymin=-2.33, ymax=1.64, 
    grid=major,
    legend pos=north east,
    legend cell align={left},
    legend style={font=\tiny, legend image post style={scale=0.8}, fill opacity=0.8,},
    xtick={-2.33, -1.64, -1.28, -0.84, 0, 0.84, 1.28, 1.64, 2.33},
    xticklabels={1\%, 5\%, 10\%, 20\%, 50\%, 80\%, 90\%, 95\%, 99\%},
    ytick={-2.33, -1.64, -1.28, -0.84, 0, 0.84, 1.28, 1.64, 2.33},
    yticklabels={1\%, 5\%, 10\%, 20\%, 50\%, 80\%, 90\%, 95\%, 99\%},
    tick label style={font=\footnotesize},
]


\addlegendimage{draw, black}
\addlegendentry{BERT}
\addlegendimage{mark=triangle, only marks, black}
\addlegendentry{Haiku}
\addlegendimage{mark=triangle*, only marks, black}
\addlegendentry{Sonnet}
\addlegendimage{mark=square, only marks, black}
\addlegendentry{gpt-4o-mini}
\addlegendimage{mark=square*, only marks, black}
\addlegendentry{o4-mini}
\addlegendimage{dashed, gray}
\addlegendentry{Chance}


\addplot[care] table[x=fpr-probit-care, y=fnr-probit-care, col sep=comma] {data/classification_MFRC_DET.csv};
\addplot[mark=triangle, only marks, care, thick] coordinates {(-0.67, -0.25)}; 
\addplot[mark=triangle*, only marks, care] coordinates {(-1.04, -0.36)}; 
\addplot[mark=square, only marks, care, thick] coordinates {(-1.08, -0.05)}; 
\addplot[mark=square*, only marks, care] coordinates {(-1.04, -0.08)}; 

\addplot[loyalty] table[x=fpr-probit-loyalty, y=fnr-probit-loyalty, col sep=comma] {data/classification_MFRC_DET.csv};
\addplot[mark=triangle, only marks, loyalty, thick] coordinates {(-1.04, 0.47)}; 
\addplot[mark=triangle*, only marks, loyalty] coordinates {(-1.34, 0.20)}; 
\addplot[mark=square, only marks, loyalty, thick] coordinates {(-1.64, 1.23)}; 
\addplot[mark=square*, only marks, loyalty] coordinates {(-1.48, 0.84)}; 

\addplot[fairness] table[x=fpr-probit-fairness, y=fnr-probit-fairness, col sep=comma] {data/classification_MFRC_DET.csv};
\addplot[mark=triangle, only marks, fairness, thick] coordinates {(-0.36, -0.20)}; 
\addplot[mark=triangle*, only marks, fairness] coordinates {(-0.67, -0.28)}; 
\addplot[mark=square, only marks, fairness, thick] coordinates {(-1.13, 0.74)}; 
\addplot[mark=square*, only marks, fairness] coordinates {(-1.04, 0.31)}; 

\addplot[authority] table[x=fpr-probit-authority, y=fnr-probit-authority, col sep=comma] {data/classification_MFRC_DET.csv};
\addplot[mark=triangle, only marks, authority, thick] coordinates {(-0.74, 0.08)}; 
\addplot[mark=triangle*, only marks, authority] coordinates {(-1.08, 0.05)}; 
\addplot[mark=square, only marks, authority, thick] coordinates {(-1.04, 0.31)}; 
\addplot[mark=square*, only marks, authority] coordinates {(-1.34, 0.74)}; 

\addplot[sanctity, thick] table[x=fpr-probit-sanctity, y=fnr-probit-sanctity, col sep=comma] {data/classification_MFRC_DET.csv};
\addplot[mark=triangle, only marks, sanctity, thick] coordinates {(-0.99, 0.55)}; 
\addplot[mark=triangle*, only marks, sanctity] coordinates {(-1.48, 0.52)}; 
\addplot[mark=square, only marks, sanctity, thick] coordinates {(-1.55, 1.28)}; 
\addplot[mark=square*, only marks, sanctity] coordinates {(-1.41, 1.08)}; 

\addplot[gray, dashed] {-x};
\end{axis}
\end{tikzpicture}
\begin{center}
\small
\textcolor{care}{\rule{0.4cm}{2pt}} Care \quad  
\textcolor{fairness}{\rule{0.4cm}{2pt}} Fairness \quad
\textcolor{loyalty}{\rule{0.4cm}{2pt}} Loyalty \quad
\textcolor{authority}{\rule{0.4cm}{2pt}} Authority \quad
\textcolor{sanctity}{\rule{0.4cm}{2pt}} Sanctity
\end{center}
\caption{Detection error tradeoff curves for MFRC dataset}
\label{fig:DET-MFRC-simple}
\end{subfigure}
\hfill
\begin{subfigure}{0.49\textwidth}
\centering
\begin{tikzpicture}
\begin{axis}[
   height=6cm,
   width=8cm,
    xlabel={False Positive Rate (Normal Deviate)},
    ylabel={False Negative Rate (Normal Deviate)},
    xmin=-2.33, xmax=1.64, 
    ymin=-2.33, ymax=1.64, 
    grid=major,
    legend pos=north east,
    legend cell align={left},
    legend style={font=\tiny, legend image post style={scale=0.8}, fill opacity=0.8,},
    xtick={-2.33, -1.64, -1.28, -0.84, 0, 0.84, 1.28, 1.64, 2.33},
    xticklabels={1\%, 5\%, 10\%, 20\%, 50\%, 80\%, 90\%, 95\%, 99\%},
    ytick={-2.33, -1.64, -1.28, -0.84, 0, 0.84, 1.28, 1.64, 2.33},
    yticklabels={1\%, 5\%, 10\%, 20\%, 50\%, 80\%, 90\%, 95\%, 99\%},
    tick label style={font=\footnotesize},
]


\addlegendimage{draw, black}
\addlegendentry{BERT}
\addlegendimage{mark=triangle, only marks, black}
\addlegendentry{Haiku}
\addlegendimage{mark=triangle*, only marks, black}
\addlegendentry{Sonnet}
\addlegendimage{mark=square, only marks, black}
\addlegendentry{gpt-4o-mini}
\addlegendimage{mark=square*, only marks, black}
\addlegendentry{o4-mini}
\addlegendimage{dashed, gray}
\addlegendentry{Chance}


\addplot[care] table[x=fpr-probit-care, y=fnr-probit-care, col sep=comma] {data/classification_MFTC_DET.csv};
\addplot[mark=triangle, only marks, care, thick] coordinates {(-0.50, -0.41)}; 
\addplot[mark=triangle*, only marks, care] coordinates {(-0.52, -0.52)}; 
\addplot[mark=square, only marks, care, thick] coordinates {(-0.67, -0.13)}; 
\addplot[mark=square*, only marks, care] coordinates {(-0.77, -0.25)}; 

\addplot[loyalty] table[x=fpr-probit-loyalty, y=fnr-probit-loyalty, col sep=comma] {data/classification_MFTC_DET.csv};
\addplot[mark=triangle, only marks, loyalty, thick] coordinates {(-1.41, 0.92)}; 
\addplot[mark=triangle*, only marks, loyalty] coordinates {(-1.48, 0.55)}; 
\addplot[mark=square, only marks, loyalty, thick] coordinates {(-1.34, 1.18)}; 
\addplot[mark=square*, only marks, loyalty] coordinates {(-1.17, 0.77)}; 

\addplot[fairness] table[x=fpr-probit-fairness, y=fnr-probit-fairness, col sep=comma] {data/classification_MFTC_DET.csv};
\addplot[mark=triangle, only marks, fairness, thick] coordinates {(-0.71, -0.41)}; 
\addplot[mark=triangle*, only marks, fairness] coordinates {(-0.84, -0.55)}; 
\addplot[mark=square, only marks, fairness, thick] coordinates {(-1.13, 0.61)}; 
\addplot[mark=square*, only marks, fairness] coordinates {(-0.88, -0.15)}; 

\addplot[authority] table[x=fpr-probit-authority, y=fnr-probit-authority, col sep=comma] {data/classification_MFTC_DET.csv};
\addplot[mark=triangle, only marks, authority, thick] coordinates {(-1.04, 0.28)}; 
\addplot[mark=triangle*, only marks, authority] coordinates {(-1.08, 0.13)}; 
\addplot[mark=square, only marks, authority, thick] coordinates {(-0.84, 0.25)}; 
\addplot[mark=square*, only marks, authority] coordinates {(-1.04, 0.52)}; 

\addplot[sanctity, thick] table[x=fpr-probit-sanctity, y=fnr-probit-sanctity, col sep=comma] {data/classification_MFRC_DET.csv};
\addplot[mark=triangle, only marks, sanctity, thick] coordinates {(-1.17, 0.71)}; 
\addplot[mark=triangle*, only marks, sanctity] coordinates {(-1.41, 0.44)}; 
\addplot[mark=square, only marks, sanctity, thick] coordinates {(-1.28, 0.99)}; 
\addplot[mark=square*, only marks, sanctity] coordinates {(-1.23, 0.77)}; 

\addplot[gray, dashed] {-x};
\end{axis}
\end{tikzpicture}
\begin{center}
\small
\textcolor{care}{\rule{0.4cm}{2pt}} Care \quad  
\textcolor{fairness}{\rule{0.4cm}{2pt}} Fairness \quad
\textcolor{loyalty}{\rule{0.4cm}{2pt}} Loyalty \quad
\textcolor{authority}{\rule{0.4cm}{2pt}} Authority \quad
\textcolor{sanctity}{\rule{0.4cm}{2pt}} Sanctity
\end{center}
\caption{Detection error tradeoff curves for MFTC dataset}
\label{fig:DET-MFTC-simple}
\end{subfigure}

\end{figure*}

%% file: main.bbl
\begin{thebibliography}{}

\bibitem[Amin et~al., 2017]{amin2017association}
Amin, A.~B., Bednarczyk, R.~A., Ray, C.~E., Melchiori, K.~J., Graham, J.,
  Huntsinger, J.~R., and Omer, S.~B. (2017).
\newblock Association of moral values with vaccine hesitancy.
\newblock {\em Nature Human Behaviour}, 1(12):873--880.

\bibitem[Bulla et~al., 2025]{bullaLargeLanguageModels2025}
Bulla, L., De~Giorgis, S., Mongiov{\`i}, M., and Gangemi, A. (2025).
\newblock Large {{Language Models}} meet moral values: {{A}} comprehensive
  assessment of moral abilities.
\newblock {\em Computers in Human Behavior Reports}, 17:100609.

\bibitem[Davis and Goadrich, 2006]{davis_relationship_2006}
Davis, J. and Goadrich, M. (2006).
\newblock The relationship between {Precision}-{Recall} and {ROC} curves.
\newblock In {\em Proceedings of the 23rd international conference on {Machine}
  learning - {ICML} '06}, pages 233--240, Pittsburgh, Pennsylvania. ACM Press.

\bibitem[Fawcett, 2006]{fawcett_introduction_2006}
Fawcett, T. (2006).
\newblock An introduction to {ROC} analysis.
\newblock {\em Pattern Recognition Letters}, 27(8):861--874.

\bibitem[Feinberg and Willer, 2013]{feinberg2013moral}
Feinberg, M. and Willer, R. (2013).
\newblock The moral roots of environmental attitudes.
\newblock {\em Psychological Science}, 24(1):56--62.

\bibitem[Forbes et~al., 2020]{forbes2020social}
Forbes, M., Hwang, J.~D., Shwartz, V., Sap, M., and Choi, Y. (2020).
\newblock Social chemistry 101: Learning to reason about social and moral
  norms.
\newblock In {\em Proceedings of the 2020 Conference on Empirical Methods in
  Natural Language Processing (EMNLP)}, pages 653--670.

\bibitem[Frimer, 2019]{Frimer_2019}
Frimer, J.~A. (2019).
\newblock Moral foundations dictionary 2.0.

\bibitem[Graham et~al., 2013]{graham2013moral}
Graham, J., Haidt, J., Koleva, S., Motyl, M., Iyer, R., Wojcik, S.~P., and
  Ditto, P.~H. (2013).
\newblock Moral foundations theory: The pragmatic validity of moral pluralism.
\newblock {\em Advances in Experimental Social Psychology}, 47:55--130.

\bibitem[Graham et~al., 2009]{graham2009liberals}
Graham, J., Haidt, J., and Nosek, B.~A. (2009).
\newblock Liberals and conservatives rely on different sets of moral
  foundations.
\newblock {\em Journal of personality and social psychology}, 96(5):1029--1046.

\bibitem[Haidt, 2012]{haidt2012righteous}
Haidt, J. (2012).
\newblock {\em The Righteous Mind: Why Good People Are Divided by Politics and
  Religion}.
\newblock Vintage.

\bibitem[Haidt and Joseph, 2004]{haidt2004intuitive}
Haidt, J. and Joseph, C. (2004).
\newblock Intuitive ethics: How innately prepared intuitions generate
  culturally variable virtues.
\newblock {\em Daedalus}, 133(4):55--66.

\bibitem[Hoover et~al., 2020]{hoover2020moral}
Hoover, J., Portillo-Wightman, G., Yeh, L., Havaldar, S., Davani, A.~M., Lin,
  Y., Kennedy, B., Atari, M., Kamel, Z., Mendlen, M., et~al. (2020).
\newblock Moral foundations twitter corpus: A collection of 35k tweets
  annotated for moral sentiment.
\newblock {\em Social Psychological and Personality Science}, 11(8):1057--1071.

\bibitem[Hopp et~al., 2021]{hopp_extended_2021}
Hopp, F.~R., Fisher, J.~T., Cornell, D., Huskey, R., and Weber, R. (2021).
\newblock The extended {Moral} {Foundations} {Dictionary} ({eMFD}):
  {Development} and applications of a crowd-sourced approach to extracting
  moral intuitions from text.
\newblock {\em Behavior Research Methods}, 53(1):232--246.

\bibitem[Kobbe et~al., 2020]{kobbe2020exploring}
Kobbe, J., Rehbein, I., Hulpu{\c{s}}, I., and Stuckenschmidt, H. (2020).
\newblock Exploring morality in argumentation.
\newblock In {\em Proceedings of the 7th Workshop on Argument Mining}, pages
  30--40.

\bibitem[Landowska et~al., 2024]{landowska2024quantitative}
Landowska, A., Budzyńska, K., and Zhang, H. (2024).
\newblock Quantitative and qualitative analysis of moral foundations in
  argumentation.
\newblock {\em Argumentation}, 38:405--434.

\bibitem[Martin et~al., 1997]{martinDETCurveAssessment1997}
Martin, A., Doddington, G., Kamm, T., Ordowski, M., and Przybocki, M. (1997).
\newblock The {{DET}} curve in assessment of detection task performance.
\newblock In {\em 5th {{European Conference}} on {{Speech Communication}} and
  {{Technology}} ({{Eurospeech}} 1997)}, pages 1895--1898. ISCA.

\bibitem[Mokhberian et~al., 2020]{mokhberian2020moral}
Mokhberian, N., Abeliuk, A., Cummings, P., and Lerman, K. (2020).
\newblock Moral framing and ideological bias of news.
\newblock In {\em Social Informatics}, pages 206--219. Springer.

\bibitem[Nguyen et~al., 2024]{nguyenMeasuringMoralDimensions2024a}
Nguyen, T.~D., Chen, Z., Carroll, N.~G., Tran, A., Klein, C., and Xie, L.
  (2024).
\newblock Measuring {{Moral Dimensions}} in {{Social Media}} with {{Mformer}}.
\newblock {\em Proceedings of the International AAAI Conference on Web and
  Social Media}, 18:1134--1147.

\bibitem[Nguyen et~al., 2022]{nguyen2022mapping}
Nguyen, T.~D., Lyall, G., Tran, A., Shin, M., Carroll, N.~G., Klein, C., and
  Xie, L. (2022).
\newblock Mapping topics in 100,000 real-life moral dilemmas.
\newblock In {\em Proceedings of the International AAAI Conference on Web and
  Social Media}, volume~16, pages 699--710.

\bibitem[Pennebaker and Francis, 1999]{pennebaker1999linguistic}
Pennebaker, J.~W. and Francis, M.~E. (1999).
\newblock {Linguistic inquiry and word count (LIWC)}.
\newblock \url{https://www.liwc.app}.
\newblock [Online; accessed 19-January-2023].

\bibitem[Preniqi et~al., 2024a]{preniqiMoralBERTFineTunedLanguage2024}
Preniqi, V., Ghinassi, I., Ive, J., Saitis, C., and Kalimeri, K. (2024a).
\newblock {{MoralBERT}}: {{A Fine-Tuned Language Model}} for {{Capturing Moral
  Values}} in {{Social Discussions}}.
\newblock In {\em Proceedings of the 2024 {{International Conference}} on
  {{Information Technology}} for {{Social Good}}}, pages 433--442, Bremen
  Germany. ACM.

\bibitem[Preniqi et~al., 2024b]{preniqi_moralbert_2024}
Preniqi, V., Ghinassi, I., Ive, J., Saitis, C., and Kalimeri, K. (2024b).
\newblock {MoralBERT}: {A} {Fine}-{Tuned} {Language} {Model} for {Capturing}
  {Moral} {Values} in {Social} {Discussions}.
\newblock In {\em Proceedings of the 2024 {International} {Conference} on
  {Information} {Technology} for {Social} {Good}}, pages 433--442, Bremen
  Germany. ACM.

\bibitem[Trager et~al., 2022]{tragerMoralFoundationsReddit2022}
Trager, J., Ziabari, A.~S., Davani, A.~M., Golazizian, P., {Karimi-Malekabadi},
  F., Omrani, A., Li, Z., Kennedy, B., Reimer, N.~K., Reyes, M., Cheng, K.,
  Wei, M., Merrifield, C., Khosravi, A., Alvarez, E., and Dehghani, M. (2022).
\newblock The {{Moral Foundations Reddit Corpus}}.

\end{thebibliography}
